\documentclass[lettersize,journal]{IEEEtran}
\usepackage{amsmath,amsfonts}
\usepackage{bm}
\usepackage{algorithmic}
\usepackage{array}
\usepackage[caption=false,font=normalsize,labelfont=sf,textfont=sf]{subfig}
\usepackage{textcomp}
\usepackage{stfloats}
\usepackage{url}
\usepackage{verbatim}
\usepackage{graphicx}
\usepackage{hyperref}
\usepackage{booktabs}
\usepackage{footnote}
\usepackage{xcolor}
\usepackage{siunitx}
\usepackage{xspace}
\usepackage[bottom]{footmisc}
\newcommand{\myrobot}{TriphiBot\xspace}

\def\BibTeX{{\rm B\kern-.05em{\sc i\kern-.025em b}\kern-.08em
    T\kern-.1667em\lower.7ex\hbox{E}\kern-.125emX}}
\usepackage{balance}
\begin{document}
\title{TriphiBot: A Triphibious Robot Combining FOC-based Propulsion with Eccentric Design
\thanks{This work was supported by China National Tobacco Corporation under Grant 110202402018. (Corresponding authors: Yanjun Cao, Junping Zhi.)}
}
\author{
    Xiangyu Li, Tiancheng Lai, Mingwei Lai, Junxiao Lin, Mengke Zhang, Junping Zhi, Chao Xu, Fei Gao and Yanjun Cao
    \thanks{
    Xiangyu Li, Tiancheng Lai, Mingwei Lai, Junxiao Lin, Mengke Zhang, Chao Xu, Fei Gao and Yanjun Cao are with the State Key Laboratory of Industrial Control Technology, Zhejiang University, Hangzhou 310027, China, and also with the Huzhou Institute of Zhejiang University, Huzhou 313000, China (email: xiangyu.li@zju.edu.cn; yanjunhi@zju.edu.cn)
    
    Junping Zhi is with Hainan Red Tower Cigarette Co., Ltd., Hainan 571100, China.}
    
    }

\markboth{}%
{Li \MakeLowercase{\textit{et al.}}: TriphiBot: A Triphibious Robot Combining FOC-based Propulsion with Eccentric Design}
\maketitle

\begin{abstract}
Triphibious robots capable of multi-domain motion and cross-domain transitions are promising to handle complex tasks across diverse environments.
However, existing designs primarily focus on dual-mode platforms, and some designs suffer from high mechanical complexity or low propulsion efficiency, which limits their application.
In this paper, we propose a novel triphibious robot with a minimalist design that combines a quadcopter structure and two passive wheels, without extra actuators.
To address the inefficiency of ground-support motion (moving on land/seabed) for quadcopter-based designs, we introduce an eccentric center of gravity (CoG) design that inherently aligns thrust with motion, enhancing efficiency without specialized mechanical designs.
Furthermore, to address the drastic differences in motion control caused by different fluids (air and water), we develop a unified propulsion system based on field-oriented control (FOC).
This method resolves torque matching issues and enables precise, rapid bidirectional thrust across different media.
Grounded in the perspective of living condition and ground support, we analyse the robot's dynamics and propose a hybrid nonlinear model predictive control (HNMPC)-PID control system to ensure stable multi-domain motion and seamless transitions.
Experimental results validate the robot's multi-domain motion and cross-mode transition capability, along with the efficiency and adaptability of the proposed propulsion system.
\end{abstract}


\section{Introduction}\label{INTRODUCTION}
\IEEEPARstart{C}{ross-domain} mobility has attracted enormous attention due to the potential to tackle complex tasks across various environments \cite{zeng2022review,ramirez2025multimodal}. 
Aerial mobility provides rapid response speed and a bird's-eye view in three-dimensional space; ground mobility features high energy efficiency, and a viewpoint from the ground; and swimming capability enables underwater inspection, thus enhancing survivability in most working environments. 
A single robot that can freely move in different conditions and provides seamless transitions between different media is highly desirable for missions in hazardous or complex areas, such as search and rescue and the exploration of unknown environments.
However, current research remains predominantly confined to dual-mode robots, such as land-air or water-air platforms \cite{kalantari2013design,li2022aerial,meng2025design}. 
While demonstrating considerable progress, these systems are inherently limited to two operational domains. 
A few studies on triphibious robots adopt a design of a quadcopter and passive wheels, which briefly demonstrate their application in three different environments \cite{zhu2019implementation,guo2018design,takahashi2015all}. 
However, they still lack the ability for seamless cross-domain movement.
\begin{figure}[t]
    \centering
    \includegraphics[width=\linewidth]{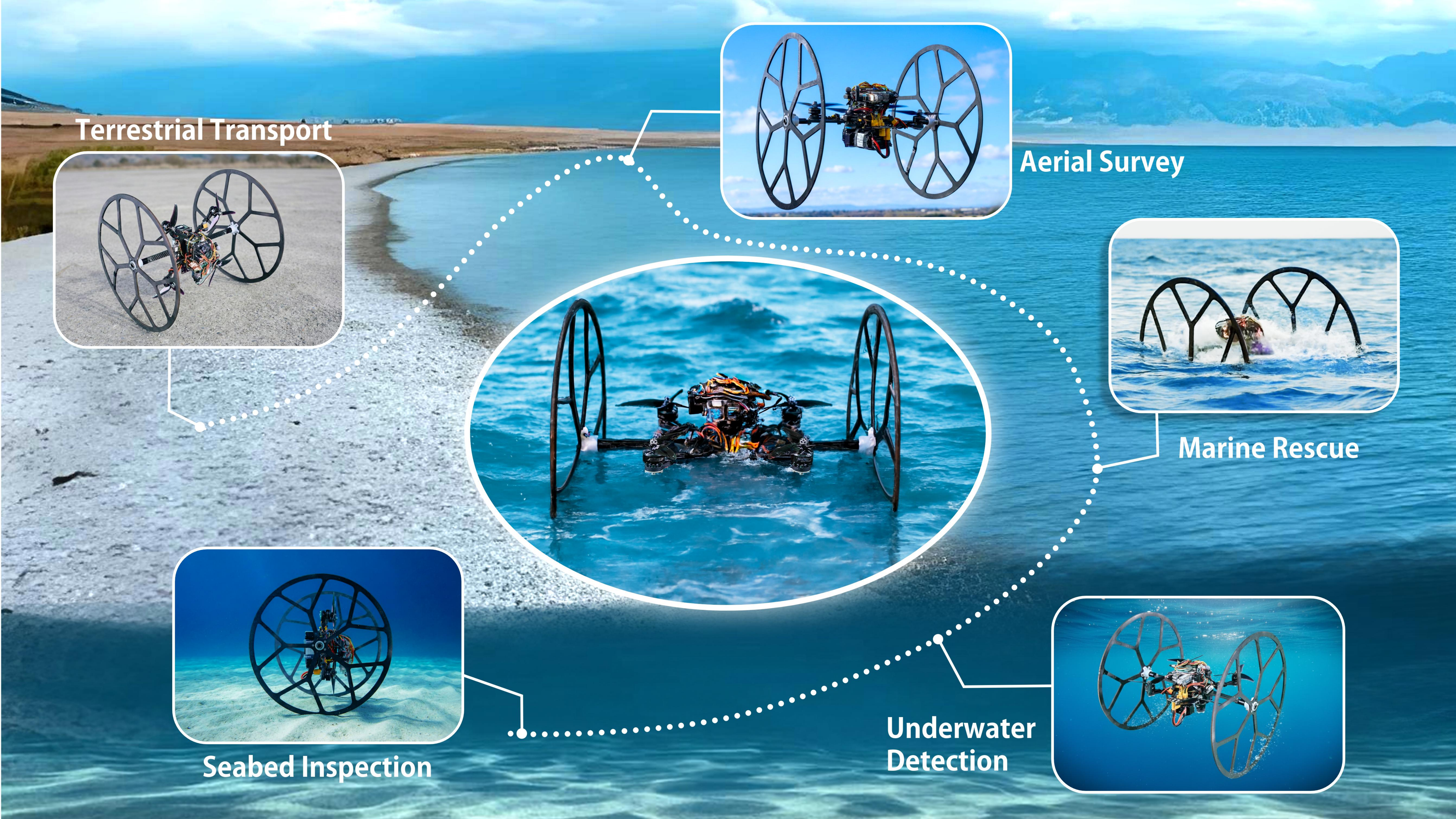}
    \caption{Depiction of the mission profile for a triphibious robot capable of flying in the air, swimming in water, and moving on land or the seabed, with smooth cross-domain, cross-mode transitions.}
    \label{depiction}
\end{figure}

The prevailing approach in cross-domain robots when involving flying is to use a multi-rotor drone as the base platform. 
These systems achieve motion in the air through the thrust generated by their rotors \cite{kalantari2013design,sun2025transverse}. 
To achieve ground motion, \cite{tan2021multimodal,xu2024flybot} rely on active wheels, which severely compromise flight efficiency due to their added drive system weight. 
Consequently, the passive-wheeled approach, which utilizes the horizontal component of rotor thrust to drive the robot's horizontal movement, has gained prominence due to its simple structure without many additional mechanisms \cite{lin2024skater,pan2023skywalker}. 
However, as only a small fraction of total thrust (horizontal component) is used to drive horizontal motion and the rest is still wasted to support the robot's weight, these passive-wheeled robots are less efficient on the ground \cite{zhang2023model}. 
To address this issue, Lai et al. \cite{lai2025trofybot} propose a passive wheeled scheme with a dynamically adjustable thrust direction through transformable mechanisms, combined with bidirectional rotor control, to obtain horizontal thrust aligned with the velocity, reducing the power consumption of ground mode. 
However, this improvement still introduces additional linkages and servo mechanisms, which not only increases the system's complexity and weight but also require additional time to complete the transition.
Robot design for water-air cross-domain motion has far fewer studies than that for air-land robots, typically relying on a vector thrust mechanism to enhance their maneuverability in water \cite{liu2023tj,sun2025transverse}.
However, these designs lack the capability to operate on the seabed or on land, focusing on the driving propulsion system challenges in the air and water. 

One of the key challenges in designing cross-domain amphibious robots comes from the different physical characteristics present in different environments, such as airflow or water flow.
Water is more than 700 times denser than air, and its viscosity is more than 50 times greater than that of air.
Considering the thrust generated by the propellers, the rotation of underwater rotors requires significantly higher torque than in air, but at a much lower rotational speed.
Some solutions employ either a compromised propulsion strategy that is suitable for air alone \cite{alzu2018loon,qin2023aerial} or two separate propulsion units specifically designed for air and water \cite{chen2020attitude,horn2019study}. 
Another category involves additional mechanisms with the same actuators to solve different torque requirements, such as integrating gearboxes \cite{liu2023tj,tan2017efficient} or utilizing deformable propellers \cite{li2022aerial,zhu2018small}.
These methods have the problem of increasing additional weight, partial underutilization of propulsion systems, and complex structures, which lead to an increase in energy consumption.
The cross-domain characteristics also place different demands on the control system.
In the air, with the large speed range of the rotation, thrust is typically achieved through open-loop or PID closed-loop control based on the electronic speed controller (ESC) \cite{lin2024skater,lai2025trofybot}.
However, due to inaccuracies in ESC speed estimation, both methods face challenges in having rapid and precise control. 
Especially during low-speed control, the speed feedback from the ESC has significant noise, leading to delays and making it difficult to switch between positive and negative thrust quickly.
This control issue becomes even more severe in aquatic environments due to the high density and viscosity of water \cite{liu2023tj,tan2017efficient}.
Therefore, developing a lightweight, unified propulsion system capable of accurately controlling the torque output in various environments is of great importance to designing a cross-domain amphibious robot.

Different from the previous work that simply considers the environment to define amphibious robots, we have found two main fundamental principles that need to be considered by analyzing the inherent requirements and motion of triphibious robots in depth.
One is the cross-domain capability from the living condition perspective (air or water), and the other is ground support from the motion feature (flying/swimming or moving on land/seabed).
The comprehensive motion strategy of the triphibious robot is depicted in Fig.~\ref{depiction}. 
Challenges come from multiple contradictory requirements and changing dynamics in different modes.
Flying in the air asks for light weight design with minimal additional mechanical structure to have a long battery life.
Underwater operation places distinct propulsion requirements, either from an extra actuator system or a specially designed mechanism to improve the thrust output.
Motion on the ground results in changing dynamics by involving ground support force and friction, both on land and seabed.
This necessitates a control system capable of rapidly adapting to varying environments and seamlessly switching strategies, ensuring stable posture, precise trajectory tracking, and smooth cross-domain mobility.
\begin{figure}[t]
    \centering
    \includegraphics[width=\linewidth]{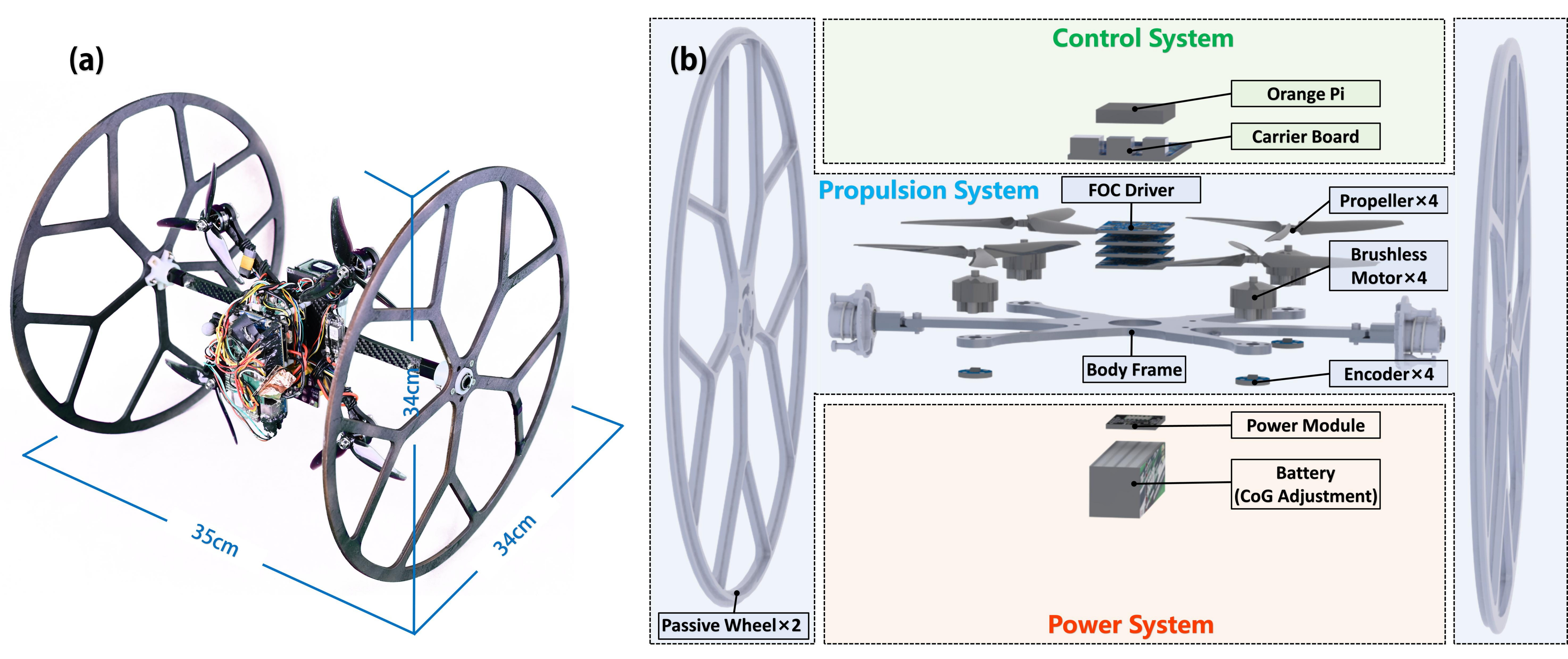}
    \caption{(a) The prototype of TriphiBot in its ground-stable state. (b) Illustration of the \myrobot's hardware layout. It can be divided into the control system, propulsion system and power system.}
    \label{boom}
\end{figure}

In this article, as shown in Fig.~\ref{boom}(a), we propose a novel triphibious cross-domain cross-mode robot following a new classification of robots' motion patterns, introducing seabed rolling mode. 
This robot firstly combines eccentric center of gravity (CoG) design and field-oriented control (FOC), overcoming the challenges of multi-medium propulsion and achieving four locomotion modes in different conditions (water, air, land, and seabed) with a simple structure.  
To the best of our knowledge, this robot is the first designed for cross-domain, cross-mode purposes and also outperforms the motion diversity of any single known creature.
The contributions are summarized as follows:

\begin{enumerate}
    \item We design and implement a novel triphibious robot that has the capability of flying in the air, swimming in water, moving on land, and on the seabed without requiring any complex mechanisms.
    \item We analyze the dynamics for different motion patterns in different medium physics and ground contact condition, providing a theoretical foundation for cross-domain cross-mode transitions.
    \item We derive the theoretical differential flatness for the robot and propose a hybrid control framework integrating hybrid nonlinear model predictive control (HNMPC) and PID for autonomous control.
    \item We combine an eccentric CoG design and an FOC propulsion system, providing an engineering implementation to improve energy efficiency and solve cross-medium torque-matching issues at the same time, backed by extensive experimental validation.
\end{enumerate}

The remainder of this article is structured as follows: Section \ref{HARDWARE DESIGN} details the design of \myrobot; Section \ref{DYNAMIC MODEL} establishes the dynamic models; Section \ref{CONTROL} presents the cross-domain control architecture; Section \ref{EXPERIMENT} provides experimental validation of multi-medium motions and transitions; and Section \ref{CONCLUSION} concludes the paper.

\begin{figure}[t]
    \centering
    \includegraphics[width=\linewidth]{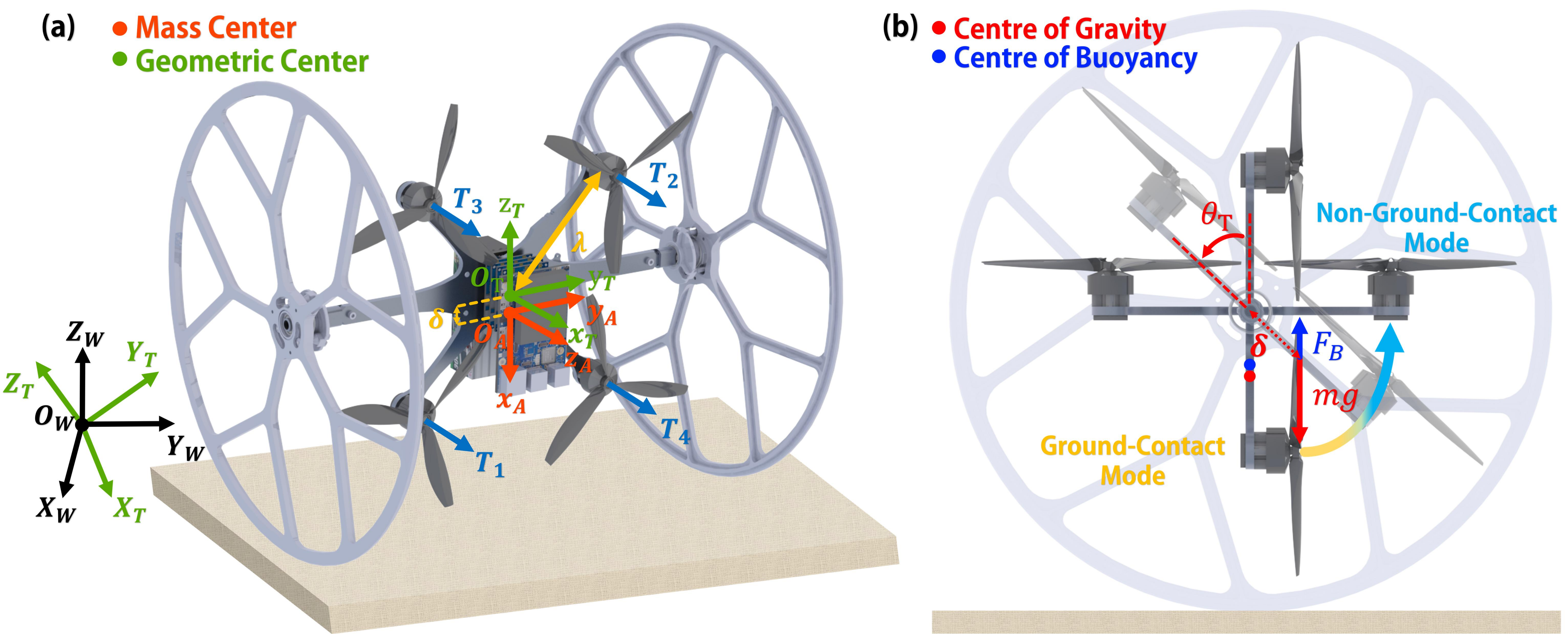}
    \caption{(a) The coordinate frames of the dynamics model of \myrobot. (b) Attitude transition of \myrobot(side view).}
    \label{frame}
\end{figure}

\section{Robot Design}\label{HARDWARE DESIGN}
\myrobot does not have many interesting features from the outlook; however, the design principle is novel by combining an eccentric design with a special propulsion system. 
\myrobot achieves an efficient triphibious motion with a straightforward but useful solution, without any additional complex mechanisms or extra actuators involved. 
\subsection{Eccentric Mass Design Principle}
As shown in Fig.~\ref{boom}(b), the robot is composed of a control system, a propulsion system and a power system.
It can be divided into a quadcopter and two passive wheels.

Fig.~\ref{frame}(a) defines three coordinate systems: the world frame ($\bm{\mathcal{F}}_{W}$, with $\mathbf{z}_{W}$ opposite to gravity), the geometric center frame ($\bm{\mathcal{F}}_{T}$ at $\mathbf{O}_{T}$), and the CoG frame ($\bm{\mathcal{F}}_{A}$ at $\mathbf{O}_{A}$). For orientation, $\mathbf{x}_{T}$ (where $\mathbf{x}_{T} \parallel \mathbf{z}_{A}$) is perpendicular to the frame plate, and $\mathbf{y}_{T}$ ($\mathbf{y}_{T} \parallel \mathbf{y}_{A}$) aligns with the wheel axis. Geometrically, $\mathbf{O}_{A}$ is shifted by $\delta$ along $-\mathbf{z}_{T}$ relative to $\mathbf{O}_{T}$.

The main novelty of \myrobot is from the principle by leveraging the gravity and combined with special propulsion system.
We specially design the position of CoG by applying the principle of the roly-poly toy. 
The CoG is located on the propeller plane with {a small shift distance} towards the front of the robot.
This design makes the main part of the drone, including the propeller plane, become vertical to the ground when the wheels touch the supporting surface, whether on land or on the seabed.
When flying through the air or swimming in water,  the small shift between the CoG and the centroid does not have much influence on performance.
Fig.~\ref{frame}(b) is the side view of robot attitude transition. 
By leveraging gravity, the robot's thrust vector is aligned with the ground motion by default, enabling full thrust to be used for propulsion.
Existing research has confirmed that robots using this method for ground movement can achieve energy savings of up to \qty{95.37}{\%} compared to aerial flight \cite{lai2025trofybot}.

Meanwhile, the restoring torque $\tau_r = mg\delta \sin\theta_T$ generated by gravity provides efficient passive pitch stabilization on land or seabed. 
When the robot takes off from the land and seabed, it controls the pitch angle until its thrust is perpendicular to the horizontal plane.
\begin{figure}[b]
    \centering
    \includegraphics[width=\linewidth]{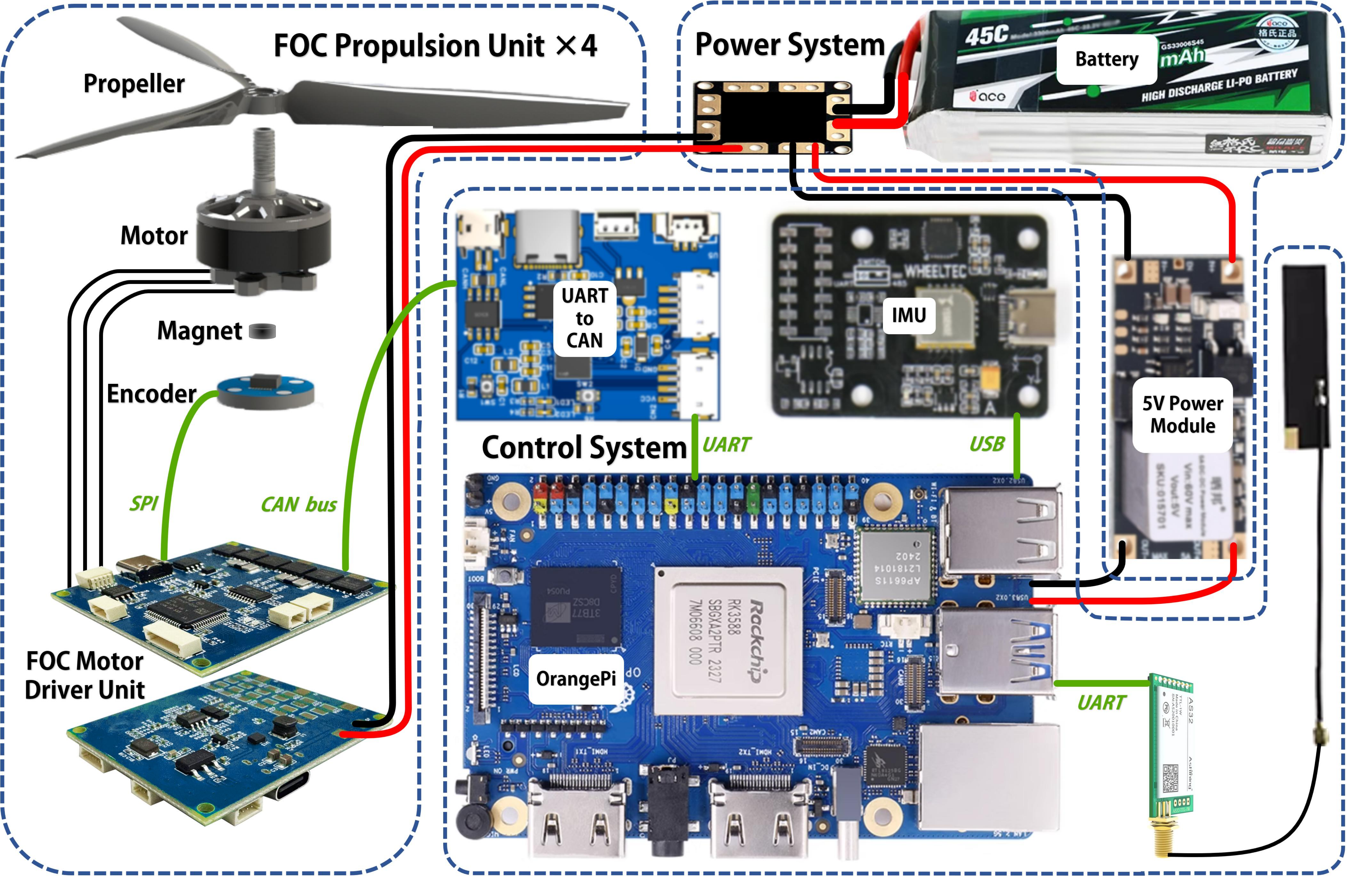}
    \caption{The electronic equipment of the \myrobot. The propulsion system consists of four FOC propulsion units.}
    \label{avionics}
\end{figure}

\myrobot is equipped with two passive wheels on both sides and four symmetrically arranged motors.
The asymmetrical layout of devices shifts the CoG along the positive $\mathbf{x}_b$ axis while maintaining symmetry along the $\mathbf{y}_b$ axis, resulting in an eccentric CoG characteristic.
The frame is sealed for waterproof and dustproof protection without compromising thermal dissipation.
The robot is designed with negative buoyancy, and its center of buoyancy (BoC) is vertically aligned slightly higher than the CoG to ensure inherent underwater stability.

\subsection{FOC based Unified Propulsion System}
The different physical characteristics and different motion modes place distinct requirement on the drive system for a triphibious robot across air, land, and water.
The motor should be capable of high torque at low speed underwater, extremely high speed in flight, and good control performance close to zero-speed rotation to generate flexible bidirectional thrust on the ground. 
It has been proven in existing literature that FOC outperforms open-loop control and estimated rotational speed feedback control in controlling the motor during high-speed rotation in the air \cite{alhaj2025optimized}.
The full-stack control architecture for triphibious motion control is described in Section \ref{CONTROL}.

Inspired by the open-source VESC\footnote{https://vesc-project.com/} project\cite{choi2020development}, we custom-designed an FOC propulsion system for small UAVs. As shown in Fig.~\ref{avionics}, the system integrates four FOC motor drive units. Our proprietary drive board features a compact, lightweight, and high-output design: weighing only \qty{12.5}{g}, it enables a high thrust-to-weight ratio, supports speeds up to \qty{200000}{ERPM} (Electrical Revolutions Per Minute), and allows rapid directional reversals.
For hardware and communication, the board utilizes an STM32F405 MCU running the VESC algorithm, communicates with the onboard computer via a CAN bus, and interfaces with an AS5047P encoder for motor position feedback. Furthermore, we update the firmware, enabling it to receive information from the onboard computer to adjust the internal control parameters.
The structural design of the \myrobot enables it to require only a relatively small forward and reverse thrust in the ground-contact mode. Under this condition, this drive unit demonstrates superior low-speed control performance compared to ESC, and can quickly and precisely output the required forward and backward thrust.

\subsection{System Electronic Design}
The electronic equipment system architecture of the robot is shown in Fig.~\ref{avionics}. 
Since different components have different power requirements, we connect the propulsion system directly to a GREPOW 4S 3300mAh lithium-ion battery. The main control board and other low-power devices are powered by a DC-DC 5V voltage regulator module.

We eliminate the need for a separate flight controller by using an interface-rich OrangePi 5 Max as the main control board, which directly reads external IMU data via serial communication.
Due to the strong attenuation of electromagnetic waves in water, we employ a dual-band communication strategy: an onboard 5GHz WiFi module for aerial operations and a 433MHz module for underwater environment.
Control instructions from the main board are transmitted via the serial port to a custom STM32F4 protocol conversion board.
After the protocol conversion, communication is established with each motor drive unit via the CAN bus.

\section{Dynamic Model}\label{DYNAMIC MODEL}
This section begins with an analysis of the dynamics of \myrobot in different motion modes, followed by a derivation of its differential flatness properties in air. 

We established a thrust and torque model for the propeller. In medium $i$ ($i \in \{a, w\}$, representing air and water, respectively), the thrust $T_i$ and torque $\tau_i$ generated by a single thruster can be expressed as:
\begin{align*}
T_i &= c_{T,i}(k_J, \text{sgn}(\varpi)) \rho_i D_p^4 \left(\frac{\varpi}{60}\right) \left|\frac{\varpi}{60}\right|\\
\tau_i &= c_{\tau,i}(k_J, \text{sgn}(\varpi)) \rho_i D_p^5 \left(\frac{\varpi}{60}\right) \left|\frac{\varpi}{60}\right|
\end{align*}
where $\rho_i$ denotes the fluid density of the current environment, and $D_p$ is the propeller diameter. $\varpi$ represents the motor rotational speed, with its sign indicating the forward or reverse rotation of the propeller. The coefficients $c_{T,i}$ and $c_{\tau,i}$ depend on the advance ratio $k_J$ ($k_J = v_s/(\varpi D_p)$), where $v_s$ is the axial relative inflow velocity. We define $d_i$ as the ratio of $c_{\tau,i}$ to $c_{T,i}$. 

\subsection{Triphibious Motion Mode Dynamics}
We categorize it into non-ground-contact mode (in air or water) and ground-contact mode (on land or seabed) based on whether \myrobot contacts the supporting surface, as shown in Fig.~\ref{frame}(b). 
In the non-ground-contact mode and the contact mode, the plane $\bm{x}_A\mathbf{O}_{A}\bm{y}_A$ is horizontal and vertical, respectively.

Coefficient $\zeta$ defines the operating medium: $\zeta=1$ for fully submerged in water, $0<\zeta<1$ for partial water contact during water-exit, and $\zeta=0$ for full detachment.
For underwater robots, added mass is the additional mass that a robot appears to have when it is accelerated or decelerated relative to the surrounding fluid.
To simplify the analysis, we define the added mass matrix and the added inertia matrix as $\bm{M}_a = \operatorname{diag}[M_{ax},M_{ay},M_{az}]$ and $\bm{I}_a = \operatorname{diag}[I_{ax},I_{ay},I_{az}]$ at the CoG, respectively.

\subsubsection{Non-Ground-Contact Mode}In the non-ground-contact mode, we conduct the analysis using the coordinate system $\bm{\mathcal{F}}_{A}$. 
The position vector in $\bm{\mathcal{F}}_{W}$ is defined as $\bm{p}_W = [p_{W,x}, p_{W,y}, p_{W,z}]^\top$, and the attitude vector is $\Theta_A = [\phi_A, \theta_A, \psi_A ]^\top$.
In coordinate system $\bm{\mathcal{F}}_{A}$, the velocity and angular rate vector are $\bm{v}_A = [v_{A,x}, v_{A,y}, v_{A,z}]^\top$ and $\bm{w}_A = [w_{A,x}, w_{A,y}, w_{A,z}]^\top$, respectively.
$\bm{R}^{W}_A$ defined as the rotation matrix that represents the transformation from $\bm{\mathcal{F}}_{A}$ to $\bm{\mathcal{F}}_{W}$ is given by
\begin{equation}
\bm{R}_{A}^{W}=\left[\begin{array}{ccc}
c \theta c \psi & s \phi s \theta c \psi-c \phi s \psi & c \phi s \theta c \psi+s \phi s \psi \\
c \theta s \psi & s \phi s \theta s \psi+c \phi c \psi & c \phi s \theta s \psi-s \phi c \psi \\
-s \theta & s \phi c \theta & c \phi c \theta
\end{array}\right],
\end{equation}

and the transformation matrix of angular rate $\bm{W}$ is given by
\begin{equation}
\bm{W}=\left[\begin{array}{ccc}
1 & s \phi t \theta & c \phi t \theta \\
0 & c \phi & -s \phi \\
0 & s \phi / c \theta & c \phi / c \theta
\end{array}\right],
\end{equation}
where $c(\cdot)$, $s(\cdot)$, and $t(\cdot)$ represent the abbreviations for $\cos(\cdot)$, $\sin(\cdot)$, and $\tan(\cdot)$, respectively. The variables $\phi$, $\theta$, and $\psi$ represent $\phi_A$, $\theta_A$, and $\psi_A$, respectively.
The \myrobot's kinematic equations of non-ground-contact mode can be expressed as
\begin{align}
    \bm{\dot{p}}_W = \bm{R}_A^W\bm{v}_A,\bm{\dot{\Theta}}_A = \bm{W} \bm{\omega}_A.
\end{align}
When \myrobot\ dives into the water, it is subjected to the buoyancy force, which is given by
\begin{equation}
    \bm{F}_B = \rho_w V\bm{g},
\end{equation}
where $\rho_w$ is the density of water, $V$ is the volume of displaced water, and $\bm{g} = [0,0,g]^\top$.
The dynamic equations are given by
\begin{align}
    &(\bm{M}_A+\zeta \bm{M}_a)\dot{\bm{v}}_A = (\bm{M}_A+\zeta \bm{M}_a)\bm{v}_A\times \bm{\omega}_A+T\bm{z}_A \notag\\
    &\hspace{7em}-{\bm{R}_A^W}^\top({m\bm{g}+\zeta\bm{F}_B})-\zeta \bm{C}_M\bm{v}_A \left\lvert\bm{v}_A\right\rvert\\
    &\hspace{7em}+ \bm{F}_{ext}\notag,\\
    &(\bm{I}_A+\zeta\bm{I}_a)\dot{\bm{\omega}}_A=\bm{\tau}_A-\bm{\omega}_A\times(\bm{I}_A+\zeta\bm{I}_a)\bm{\omega}_A \notag\\
    &\hspace{7em}-\bm{P}_B^A\times{\bm{R}_A^W}^\top\bm{F}_B+(\zeta\bm{I}_a\bm{v}_A)\times\bm{v}_A\notag\\
    &\hspace{7em}-\zeta \bm{C}_I\bm{\omega}_A\left\lvert \bm{\omega}_A \right\rvert+\bm{\tau}_{ext}
\end{align}
where $\bm{M}_A = \operatorname{diag}[m,m,m]$ and $\bm{I}_A = \operatorname{diag}[I_{Ax},I_{Ay},I_{Az}]$ are the mass and inertia matrices defined at the CoG, respectively. $m$ is the total mass of the robot. $\times$ is the cross product and $T$ is collective rotor thrust. 
$\bm{C}_M$ and $\bm{C}_I$ are the drag coefficient matrix and the drag torque coefficient matrix in water, respectively, given by $\bm{C}_M = \operatorname{diag}[C_{u|u|}, C_{v|v|}, C_{w|w|}]$ and $\bm{C}_I = \operatorname{diag}[C_{p|p|}, C_{q|q|}, C_{r|r|}]$.
$\bm{F}_{ext}$ and $\bm{\tau}_{ext}$ are the external force and torque, respectively. $\bm{P}_B^A = [-p_{b,A}, 0, 0]^\top$ is the position vector from the CoG to the buoyancy center.
$\bm{\tau}_A = [\tau_{A,x}, \tau_{A,y}, \tau_{A,z}]^\top$ is control torque, and because of its asymmetric design, it can be expressed as
\begin{equation}\label{thrust_torque_a}
\begin{bmatrix}T\\\tau_{A,x} \\\tau_{A,y} \\\tau_{A,z}\end{bmatrix} =  
\begin{bmatrix}1 & 1 & 1 & 1 \\
        -\frac{\lambda}{\sqrt{2}} & \frac{\lambda}{\sqrt{2}}  & \frac{\lambda}{\sqrt{2}} & -\frac{\lambda}{\sqrt{2}}\\
        \delta-\frac{\lambda}{\sqrt{2}} & \frac{\lambda}{\sqrt{2}}+\delta & \delta-\frac{\lambda}{\sqrt{2}} & \frac{\lambda}{\sqrt{2}}+\delta\\
        -d_i & -d_i & d_i& d_i\\
\end{bmatrix}\bm{u},
\end{equation}
where $\lambda$ is the arm length, $\delta$ is the distance from the centroid to the CoG and $\bm{u}=[T_1,T_2,T_3,T_4]^\top$ is the thrust of each motor.

\subsubsection{Ground-Contact Mode}In the ground-contact mode, we assume that the contact surface is flat and there is no lateral movement of the wheels. 
For the convenience of analysis, we conduct the analysis using the coordinate system $\bm{\mathcal{F}}_{T}$.
$\bm{\mathcal{F}}_{T}$ is obtained by rotating $\bm{\mathcal{F}}_{A}$ 90 degrees in the positive direction of $\bm{y}_{A}$.
The attitude and angular rate vector are $\bm{\Theta}_T=[\phi_T ,\theta_T,\psi_T ]^\top$ and $\bm{w}_T = [w_{T,x}, w_{T,y}, w_{T,z}]^\top$, respectively. 
The kinematic equations of ground-contact mode are given by
\begin{align}
&\bm{\dot{p}}_{W} = [v_l\cos\psi_T, v_l\sin\psi_T,0]^\top,\\
&\dot{v}_l = \frac{T\cos\theta_T-\zeta C_{w|w|}v_l^2}{(m+\zeta M_{az})}, \label{v_l_1}
\end{align}
where $v_l$ is a scalar quantity representing the speed on a horizontal plane.
The dynamic equations are given by
\begin{align}
&\bm{\dot\Theta}_T = [\dot{\phi}_T,\dot{\theta}_T,\dot{\psi}_T]^\top = [0,w_{T,y},w_{T,z}\sec\theta_T]^\top ,\\
&\bm{\dot{\omega}}_T=(\bm{I}_T+\zeta\bm{I}_a)^{-1}[\bm{\tau}_T-\bm{\omega}_T\times(\bm{I}_T+\zeta\bm{I}_a)\bm{\omega}_T \notag  \\
&\hspace{2.5em}-\bm{\tau}_g-\zeta(\bm{P}_B^T\times{\bm{R}_T^W}^\top\bm{F}_B+\bm{C}_I\bm{\omega}_T\left\lvert \bm{\omega}_T \right\rvert)]\label{dot_omega},
\end{align}
where $\bm{I}_T = \operatorname{diag}[I_{Tx},I_{Ty},I_{Tz}]$ represents the moment of inertia tensor defined at the geometric center, $\bm{\tau}_{g} = [0, m\|\bm{g}\|\delta\sin\theta_T, 0]^\top$ is the gravity moment vector, and $\bm{P}_B^T = [0, 0, p_{b,T}]^\top$. $\bm{\tau}_T = [\tau_{T,x}, \tau_{T,y}, \tau_{T,z}]^\top$ is the control moment vector and it can be expressed as
\begin{equation} \label{thrust_torque_t}
\begin{bmatrix}
        T \\ \tau_{T,x} \\ \tau_{T,y} \\ \tau_{T,z}
\end{bmatrix} =  
\begin{bmatrix}1 & 1 & 1 & 1 \\
        -\frac{\lambda}{\sqrt{2}} & \frac{\lambda}{\sqrt{2}}  & \frac{\lambda}{\sqrt{2}} & -\frac{\lambda}{\sqrt{2}}\\
        -\frac{\lambda}{\sqrt{2}} & \frac{\lambda}{\sqrt{2}} & -\frac{\lambda}{\sqrt{2}} & \frac{\lambda}{\sqrt{2}}\\
        -d_i & -d_i & d_i & d_i\\
\end{bmatrix}\bm{u}.
\end{equation}
To sum up, the state vector of the hybrid system is $\bm{X} = [\bm{p}_W^\top, \bm{v}_A^\top, \bm{\Theta}_A^\top, \theta_T, \psi_T, \bm{w}_A^\top, \bm{w}_T^\top,v_l]^\top$ and the input vector is $\bm{u} = [T_1, T_2, T_3, T_4]^\top$.

\subsection{Differential Flatness}
When the robot is in aerial non-ground-contact mode, based on the analysis of the CoG, it can be regarded as a regular quadcopter. 
The differential flatness of the quadcopter in the aerial mode has been well studied in previous works \cite{mellinger2011minimum,faessler2017differential,watterson2019control,wang2022robust}, where the flat output is selected as $ \bm{\sigma}_f = [p_{W,x}, p_{W,y}, p_{W,z}, \psi_A]^\top$. 
The state and input vectors are $\bm{X}_a = [\bm{p}_W^\top,\bm{v}_A^\top,\bm{\Theta}_W^\top, \bm{\omega}_A^\top]^\top$ and $\bm{u} = [T_1, T_2, T_3, T_4]^\top$, respectively. 

When the robot contacts the ground, we define the flat output of terrestrial mode as $\bm{\sigma}_t = [p_{W,x},p_{W,y},\theta_T]^\top$. 
In order to prove the differential flatness on the land, we need to derive each state and input in $\bm{X}_t = [\bm{p}_W^\top,v_l,\theta_T, \psi_T, \bm{\omega}_T^\top]^\top$ and $\bm{u} = [T_1, T_2, T_3, T_4]^\top$ from $\bm{\sigma}_t$ and its derivatives. 
As we assume that the robot is not skidding on land, the $\psi_T$ value is given as 
\begin{equation}
\psi_T  = \operatorname{arctan2}(\kappa \dot{p}_{W, y}, \kappa \dot{p}_{W, x}),
\end{equation}
where $\kappa$ represents the direction of the robot's movement (when $\kappa = 1$, it moves forward, when $\kappa = -1$, it moves backward). We can directly get $\bm{p}_W$ and $\theta_T$ from $\bm{\sigma}_t$ and we do not want it to rotate around the $\bm{x}_T$. Therefore, we expect $\tau_{T,x} = 0$ and we can derive $\bm{\omega}_T$ and $\bm{\dot\omega}_T$ from $\theta_T$ and $\psi_T$.
\begin{align} \label{omega_dot_omega}
\bm{\omega}_T=
\begin{bmatrix}
    -\sin\theta_T \dot{\psi}_T \\
    \dot{\theta}_T \\
    \dot{\psi}_T \cos \theta_T
\end{bmatrix},
\bm{\dot\omega}_T=
\begin{bmatrix}
    -\cos\theta_T \dot{\theta}_T\dot{\psi}_T-\sin\theta_T \ddot{\psi}_T  \\
    \ddot{\theta}_T \\
    \ddot{\psi}_T \cos \theta_T-\dot{\psi}_T\dot{\theta}_T \sin \theta_T
\end{bmatrix}.
\end{align}
$v_l$ is given by
\begin{equation}\label{v_l_2}
    v_l = \kappa \sqrt{(\dot{p}_{W,x})^2+(\dot{p}_{W,y})^2}.
\end{equation}
To maximize energy efficiency, we keep $x_T$ parallel to the land, namely $\theta_T = 0$. By substituting (\ref{v_l_2}) into (\ref{v_l_1}), the total thrust $T$ is given by
\begin{equation}
    T = m\dot{v}_l=\frac{m(\ddot{p}_{W,x}\dot{p}_{W,x}+\ddot{p}_{W,y}\dot{p}_{W,y})}{\cos\theta_T\sqrt{(\dot{p}_{W,x})^2+(\dot{p}_{W,y})^2}}.
\end{equation}
Then by combining (\ref{dot_omega}),(\ref{omega_dot_omega}), we can get $\bm{\tau}_{T}$.
Finally, we obtain $\bm{u}$ with $T$ and $\bm{\tau}_T$ substituted into the inverse form of (\ref{thrust_torque_t}).
All the variables of state $\bm{X}_t = [\bm{p}_W^\top,v_l,\theta_T, \psi_T, \bm{\omega}_T^\top]^\top$ and input $\bm{u}=[T_1,T_2,T_3,T_4]^\top$ can be obtained from the flat output. Therefore, we can conclude that the terrestrial dynamics of \myrobot is differentially flat.
\begin{figure}[b]
    \centering
    \includegraphics[width=\linewidth]{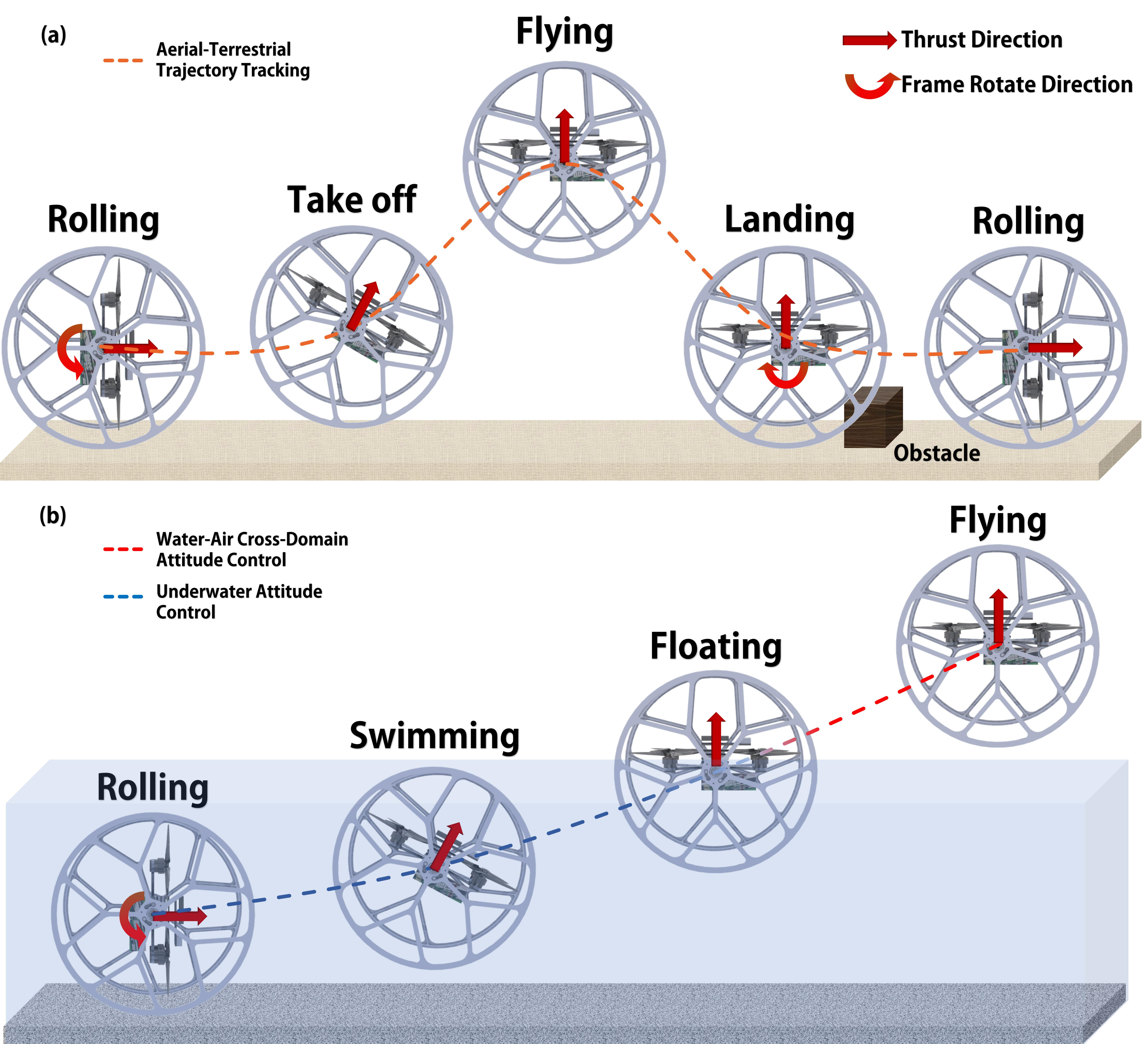}
    \caption{Conceptual diagram of cross-domain transitions. (a) After changing the body's attitude, robot takes off, flies and lands, including crossing obstacles by changing the attitude. (b) From the seabed to the air.}
    \label{modchange}
\end{figure}
\section{Control Architecture and Motor Control}\label{CONTROL}
This section presents \myrobot's cross-domain control framework and mode-switching mechanism (Fig.~\ref{modchange}). Furthermore, it details the motor's FOC method for automatic speed and torque adaptation.

\subsection{Hybrid Control Architecture}
\begin{figure}[t]
    \centering
    \includegraphics[width=\linewidth]{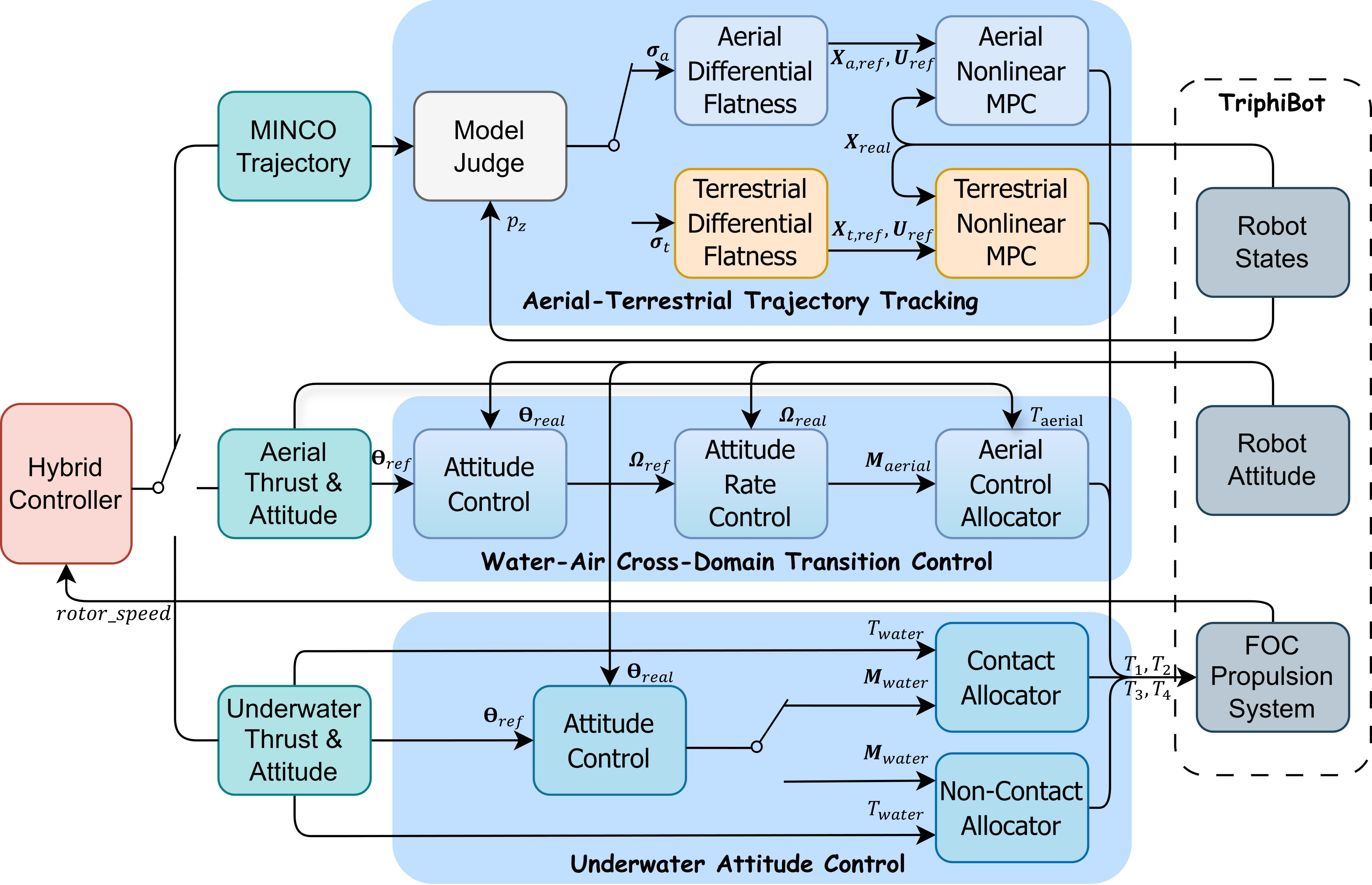}
    \caption{The hybrid control structure of \myrobot. The controller automatically determines the switching variable $\eta$ based on the reference height $p_z$. The transitions are accomplished within a unified HNMPC framework. When transferring from water to the air or the land, it needs to be controlled through PID methods.}
    \label{ctr}
\end{figure}
Since high-rate communication and precise positioning are accessible in the air but severely limited in underwater environments and during air-water transition phases, we developed distinct control methods that are activated according to the operation mode.
As shown in Fig.~\ref{ctr}, we design a HNMPC-PID control system based on a finite state machine.

\noindent\textbf{Aerial-Terrestrial Trajectory Tracking.} For both aerial and terrestrial position control, we obtain a flat-output trajectory from MINCO \cite{wang2022geometrically}, then derive the reference trajectory utilizing the differential flatness.
We design an HNMPC controller, which accurately distinguishes the dynamics of the robot by using discrete switching variable $\eta$. 
The controller automatically determines the switching variable $\eta$ based on the reference $p_z$.

The HNMPC finds the best control commands by solving an optimal control problem (OCP) in a receding horizon manner. 
It aims to minimize the cost function of the error between the predicted state of measurement and the reference trajectory within the time range $[t_0, t_0 + h]$. 
The time range is $h=N\cdot dt$, where $dt$ represents the time step and $N$ is the number of steps.

The aerial and terrestrial dynamics of the system are described as 
\begin{equation}
f(\bm{X}, \bm{u}) = \eta f(\bm{X}_t, \bm{u} ) + (1 - \eta)f(\bm{X}_a, \bm{u}).
\end{equation}
Then, the NMPC problem is transformed into a nonlinear programming problem, and the optimal control command sequence $\bm{u}^*\in \mathbb{R}^{4\times N}$ is generated as
\begin{equation}
\begin{aligned}
\bm{u}^*  = &\arg\min_{\bm{u}} \sum_{i = k}^{k+N-1}\left[\eta \cdot \tilde{\bm{X}}_{t, i}^\top \bm{Q}_{t} \tilde{\bm{X}}_{t, i}\right. \\
& \left.+\left(1-\eta\right) \cdot \tilde{\bm{X}}_{a, i}^\top \bm{Q}_{a} \tilde{\bm{X}}_{a, i}+\tilde{\bm{u}}_{i}^\top \bm{Q}_{u} \tilde{\bm{u}}_{i}\right] \\
& +\eta \cdot \tilde{\bm{X}}_{t, k+N}^\top \bm{Q}_{t} \tilde{\bm{X}}_{t, k+N}+\left(1-\eta\right) \tilde{\bm{X}}_{a, k+N}^\top \bm{Q}_{a} \tilde{\bm{X}}_{a, k+N},\\
& \text{s.t.}\hspace{0.5em}\bm{X}_{i+1}=f(\bm{X}_{i},\bm{u}_{i}), \bm{u}_{i} \in [\bm{u}_{min},\bm{u}_{max}],
\end{aligned}
\end{equation}
where $\eta$ is the discrete variable, which is given by
\begin{equation}
\eta =\begin{cases}
1,\text{if $p_{W,z,k}<h_{\text{judge}}$}  \\
0,\text{if $p_{W,z,k}>h_{\text{judge}}$}
\end{cases},
\end{equation}
where $h_{\text{judge}}$ is the threshold used to determine whether to take off. $k$ is the current time step, $\tilde{\bm{X}}_{t, i} = \bm{X}_{t,ref}-\bm{X}_{t,i}, \tilde{\bm{X}}_{a, i} = \bm{X}_{a,ref}-\bm{X}_{a,i}$ and $\tilde{\bm{u}}_i = \bm{u}_{ref,i}-\bm{u}_i$ are the terrestrial states, aerial states and inputs errors, respectively. $\tilde{\bm{X}}_{t, k+N}$ and $\tilde{\bm{X}}_{a, k+N}$ are the end states error. $\bm{Q}_{t},\bm{Q}_{a}$ and $\bm{Q}_u$ are the terrestrial state, aerial state and input weight matrices written as
\begin{align}
\boldsymbol{Q}_{t} & = \operatorname{diag}\left(\left[\boldsymbol{Q_p}, \boldsymbol{Q_w}, Q_{v_{l}}, Q_{\theta}, Q_{\psi}\right]\right), \\
\boldsymbol{Q}_{a} & = \operatorname{diag}\left(\left[\boldsymbol{Q_p}, \boldsymbol{Q_v}, \boldsymbol{Q_\Theta}, \boldsymbol{Q_w}\right]\right), \\
\boldsymbol{Q}_u & = \operatorname{diag}\left(\left[Q_{T_1},Q_{T_2},Q_{T_3},Q_{T_4}\right]\right) .
\end{align}

\noindent\textbf{Water-Air Cross-Domain Transition Control.}
Since the positioning of the robot from water to air cannot be seamless, and waves can affect the stability of positioning as the robot floats on the water, we use cascade PID control for cross-domain transition control when it jumps out of the water.
In the aerial attitude controller, the inner loop tracks angular velocity, while the outer loop tracks angle.
Then, the control allocator maps the required torque into the required thrust $\bm{u}$ according to (\ref{thrust_torque_a}).

When the robot jumps out of the water, the rotational speed of rotors increases to tens of thousands of RPM (Revolutions Per Minute).
Conversely, upon water entry, the speed drops rapidly to a few hundred RPM.
The hybrid controller determines whether the robot has left the water surface by monitoring the changes in the rotor speed, and then triggers the corresponding controller.
When the robot is out of the water and its attitude is stable, it can switch to trajectory tracking mode.

\noindent\textbf{Underwater Attitude Control.} 
Due to underwater localization and communication constraints, we implemented a single-loop PID attitude control system. The controller utilizes the angular tracking error to generate a three-axis corrective torque, which is then mapped alongside the total thrust command from the remote controller to determine the individual rotational speeds of the four motors.

\subsection{FOC Motor Control}

The ESCs rely on back-electromotive force (back-EMF) for six-step square wave commutation.
This method fails to keep the stator magnetic field perfectly perpendicular to the rotor flux, leading to suboptimal efficiency.
Furthermore, weak back-EMF signals at low speeds often cause commutation errors, oscillations, and poor low-speed performance.

\begin{figure}[t]
    \centering
    \includegraphics[width=\linewidth]{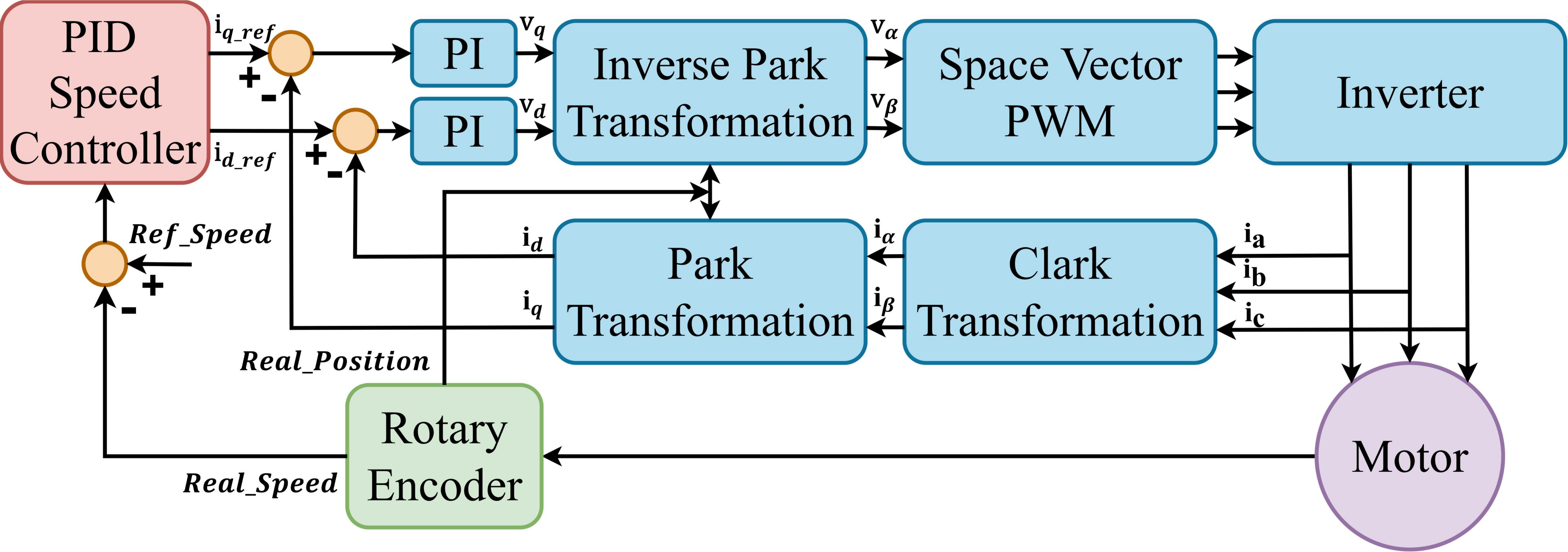}
    \caption{The FOC algorithm, in which $\text{i}_{q\_ref}$ determines the motor torque. This algorithm adjusts the current $\text{i}_{q\_ref}$ based on the rotational speed error to achieve torque matching. The current $\text{i}_{d\_ref}$ is usually set to 0.}
    \label{FOC Algorithm}
\end{figure}

To overcome these limitations, we adopt FOC equipped with an encoder for precise rotor angle ($\theta_m$) feedback. As illustrated in Fig.~\ref{FOC Algorithm}, the FOC system employs a cascaded PID structure comprising an outer speed loop and an inner current loop~\cite{BPRA073}. The nonlinear AC stator currents ($\text{i}_a, \text{i}_b, \text{i}_c$) are mapped into a linear DC control problem in the synchronous $d$-$q$ frame via sequential Clarke and Park transformations:
\begin{equation}
\begin{aligned}
    \begin{bmatrix} \text{i}_\alpha \\ \text{i}_\beta \end{bmatrix} &= \frac{2}{3}
    \begin{bmatrix} 1 & -\frac{1}{2} & -\frac{1}{2} \\ 0 & \frac{\sqrt{3}}{2} & -\frac{\sqrt{3}}{2} \end{bmatrix}
    \begin{bmatrix} \text{i}_a \\ \text{i}_b \\ \text{i}_c \end{bmatrix}, \\
    \begin{bmatrix} \text{i}_d \\ \text{i}_q \end{bmatrix} &= 
    \begin{bmatrix} \cos\theta_m & \sin\theta_m \\ -\sin\theta_m & \cos\theta_m \end{bmatrix}
    \begin{bmatrix} \text{i}_\alpha \\ \text{i}_\beta \end{bmatrix}.
\end{aligned}
\end{equation}

By operating in this decoupled frame, the stator magnetic field is actively maintained perpendicular to the rotor's permanent magnetic field, ensuring maximum torque generation per ampere.
Notably, when the robot operates underwater and immense fluid drag prevents the motor from reaching the target speed, the outer PID speed loop automatically increases the reference current $\text{i}_{q\_ref}$ to achieve adaptive torque matching.
Building upon this, we dynamically adjust the PID parameters based on the robot's operating environment.
Specifically, ground rolling and submerged cruising require rapid command tracking, thus utilizing high-response parameters.
Conversely, during aerial flight, the system switches to a smoother parameter configuration to ensure attitude stability.

\section{Experiment}\label{EXPERIMENT}

In this section, we set up a test bench to evaluate the propulsion capability of the propulsion unit through static output characteristics.
Meanwhile, we conduct a comprehensive validation of the \myrobot's motion and cross-domain capability through trajectory tracking, attitude control, and cross-domain transition experiments across aerial, terrestrial and aquatic environments.

\subsection{Experiments Setup}
The physical parameters and controller parameters are listed in Table \ref{parameters}. 
The global localization when testing with controller is obtained from the NOKOV\footnote{https://www.nokov.com/} motion capture system. 
We solve the NMPC problem using ACADO toolkit \cite{houska2011acado} and qpOASES \cite{ferreau2014qpoases}, with a control frequency of 200Hz. 
We use the root-mean-square error to evaluate the tracking performance, 
which is expressed as \( \text{RMSE} = ( \tfrac{1}{N} \sum_{k=1}^{N}\| \mathbf{p}^k - \mathbf{p}^k_{\text{ref}} \|^2 )^{1/2} \).
Among them, $\bf{p}^k$ and $\bf{p}^k_{ref}$ represent the actual and reference positions sampled at the $k$-th step respectively.

\begin{table}[b]
\centering
\caption{System Parameters and Control Weights}\label{parameters}
\resizebox{\columnwidth}{!}{
\begin{tabular}{lc|lc}
\toprule
Parameter & Value & Parameter & Value \\  
\midrule
Size ($L\times W\times H$) [$\mathrm{cm}$]   & $35\times 34\times 34$       & $N, dt$ [$\mathrm{s}$]      & $40, 0.05$\\
$m$ [$\mathrm{kg}$], $g$ [$\mathrm{m/s^2}$]  & $1.1, 9.8$                   & $\bm{Q_p}$   & $\operatorname{diag}(5000,5000,3000)$\\
$\bm{I}_{A}$ [$\mathrm{kg\cdot m^2}$]        & $\operatorname{diag}(0.0113,0.0018,0.0125)$ & $\bm{Q_v}$   & $\operatorname{diag}(500,500,500)$\\
$\bm{I}_{T}$ [$\mathrm{kg\cdot m^2}$]        & $\operatorname{diag}(0.0125,0.0018,0.0113)$ & $\bm{Q_\Theta}$& $\operatorname{diag}(500,500,500,500)$\\
$c_{\tau,a}, c_{\tau,w}$                     & $0.0201, 0.0287$             & $\bm{Q_{w}}$ & $\operatorname{diag}(10,10,10)$\\
$c_{T,a,f}, c_{T,a,r}$                       & $0.163, 0.079$               & $\bm{Q_u}$   & $\operatorname{diag}(100,100,100,100)$ \\
$c_{T,w,f}, c_{T,w,r}$                       & $0.234, 0.111$               & $Q_{v_l}, Q_\theta, Q_\psi$ & $500, 500, 500$ \\
$\lambda, \delta, D_p$ [$\mathrm{cm}$]       & $8, 1.5, 12.954$             & $\bm{C}_M$ [$\mathrm{N/(m/s)^2}$] & $\operatorname{diag}(15.7,15.0,16.4)$ \\
$\rho_a, \rho_w$ [$\mathrm{kg/m^3}$]         & $1.225, 1000$                & $\bm{C}_I$ [$\mathrm{N\cdot m/(rad/s)^2}$] & $\operatorname{diag}(0.31,0.26,0.17)$ \\
\bottomrule
\end{tabular}}
\end{table}

\begin{figure}[t]
    \centering
    \includegraphics[width=\linewidth]{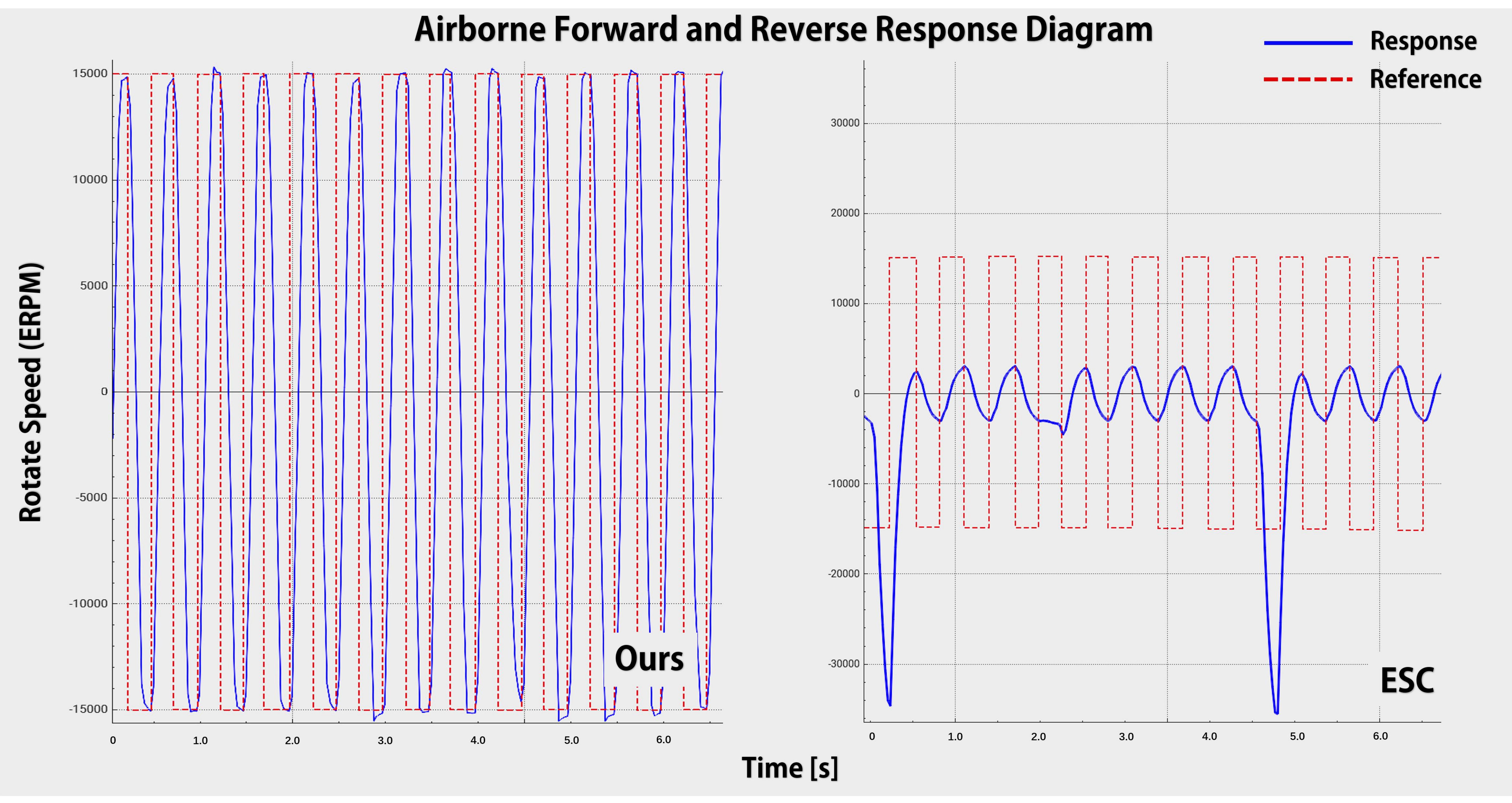}
    \caption{The response of the rotor to high-speed forward and backward rotation in air using different drivers.}
    \label{response}
\end{figure}
\subsection{Output Characteristics of the Propulsion System}

We validate the proposed propulsion unit's performance through a series of tests in this section.

\noindent\textbf{Rapid forward/backward thrust switching.} First, a rapid forward/backward thrust switching experiment is conducted in the air. 
We adopt a closed-loop speed control using a ESC with the same motor and propeller for comparison.
The speed command undergoes a step change within the range of $\pm$\qty{15000}{ERPM} with a directional switching interval of \qty{0.25}{s}, and the experimental results are shown in Fig.~\ref{response}.

Under such a high-frequency switching requirement, the ESC cannot follow the command correctly. However, the response curve of the proposed propulsion unit is smooth and no overshoot phenomenon is observed.
This result proves that our propulsion unit is superior to ESC and can meet the requirement for rapid switching between forward and reverse thrusts.

\noindent\textbf{Fast startup and precise speed control underwater.} 
Fig.~\ref{driver_test}(a) shows the dedicated experimental platform established for the performance test of the propulsion unit.
Fig.~\ref{driver_test}(b) presents the comparative experimental results of the underwater startup process of the motor. 
We set the underwater target speed at \qty{420}{RPM}, and it can be observed that the motor controlled by the ESC exhibits oscillation during the startup stage and has obvious steady-state error. In contrast, the proposed propulsion unit shows better dynamic performance. It reaches \qty{90}{\%} of the expected speed in approximately \qty{0.14}{s}, with a smooth and rapid transition. Its performance underwater is superior to that of ESC.

\noindent\textbf{Underwater force-speed mapping relationship.} We test the mapping relationship between the rotational speed and thrust of different propellers (DJI7455$\times$3 and 51477R$\times$3). 
The experimental data shown in Fig.~\ref{driver_test}(b) indicate that, under the same thrust, the rotational speed of the propellers underwater (in both directions) is lower than the measured values in the air.

\begin{figure}[t]
    \centering
    \includegraphics[width=\linewidth]{figures/driver_test.pdf}
    \caption{The static performance testing of the propulsion system: (a) Test Platform. (b) The underwater step response with different drivers. (c) The underwater force-rotation speed curves of different propellers. (d) The underwater specific thrust-rotation speed curves compared with TJ-FlyingFish. (e) The underwater power-rotation speed curves of different drivers. (f) The underwater specific thrust-rotation speed curves of different drivers.}
    \label{driver_test}
\end{figure}
\begin{figure*}[t]
    \centering
    \includegraphics[width=\linewidth]{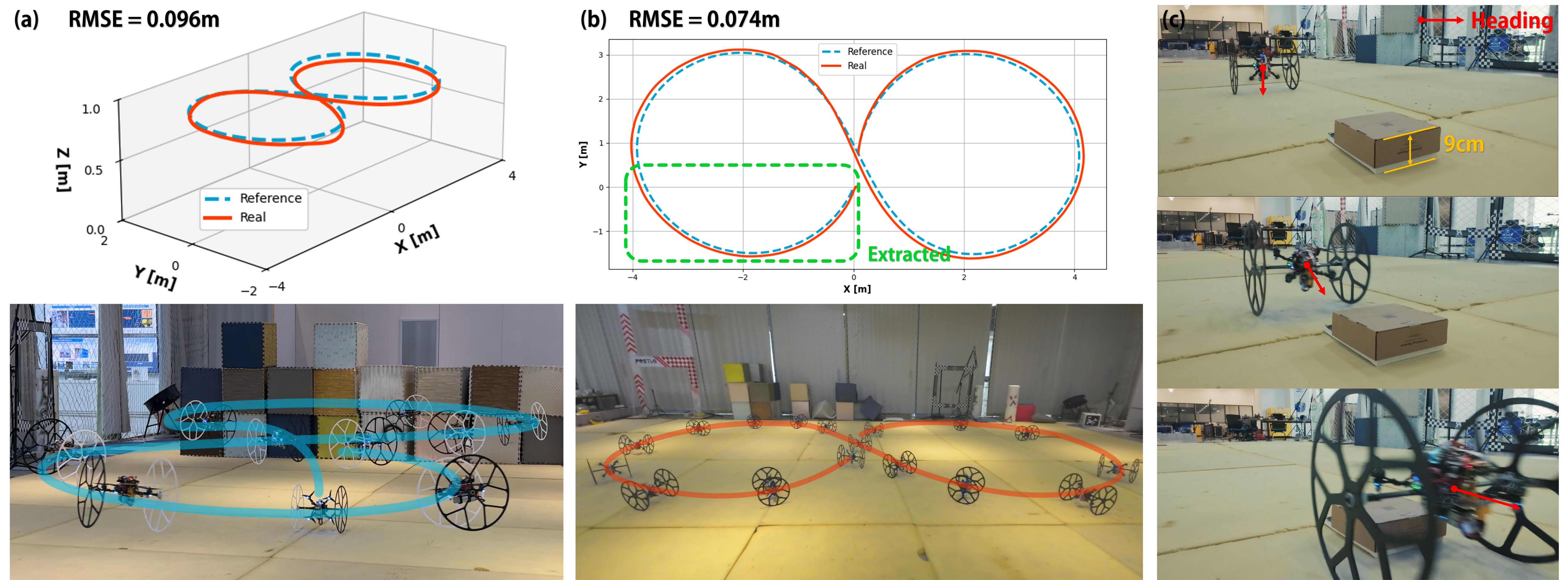}
    \caption{(a) Reference and real $\bm{p}_{W,xyz}$ when tracking an 8-shaped aerial trajectory. (b) Reference and real $\bm{p}_{W,xy}$ when tracking an 8-shaped terrestrial trajectory, the green dashed rectangular box indicates the extracted section of the trajectory. (c) By changing the pitch angle to overcome ground obstacles.} 
    \label{rmse}
\end{figure*}
\noindent\textbf{Underwater high-efficiency verification.} When the torque output of a brushless motor is inadequate, torque matching fails, and the efficiency decreases. To quantitatively evaluate the superiority of the proposed propulsion unit in terms of underwater efficiency over the mechanical torque matching device and ESC, we conduct power consumption and specific thrust comparison experiments at different rotational speeds.
As shown in Fig.~\ref{driver_test}(d), when using the same motor (T-MOTOR AT2312-KV1150) and rotor (DJI9455$\times$3), our propulsion unit is superior to TJ-FlyingFish~\cite{liu2023tj} which is equipped with a mechanical torque matching device. The maximum specific thrust of TJ-FlyingFish is \qty{0.265}{N/W}, whereas ours reaches \qty{0.898}{N/W} under the same conditions.
In addition, as shown in Fig.~\ref{driver_test}(e) and (f), within the entire testing speed range, the power consumption of the ESC is much higher than that of our unit. Specifically, the ESC's maximum power consumption reaches up to five times that of our unit. Our propulsion unit achieves a value of up to \qty{0.825}{N/W}, a performance level 14 times greater than the traditional ESC's \qty{0.059}{N/W}. The test results further show that the ESC fails to provide adequate torque for the underwater load due to its control strategy, thus operating at extremely low efficiency. This limitation restricts its maximum speed to only \qty{700}{RPM} under the maximum allowable continuous current.

\subsection{Air Domain Experiments}
\noindent\textbf{Non-ground-contact mode.} Since \myrobot  is affected by deflection torque, model accuracy significantly impacts flight performance.
Therefore, we evaluate the maneuverability of \myrobot through takeoff and 8-shaped trajectory tracking experiments.
The robot performed this task at a maximum speed of \qty{2}{m/s} and acceleration of \qty{2}{m/s^2}.
The trajectory tracking result is shown in Fig.~\ref{rmse}(a), with RMSE of \qty{0.096}{m}.
This experiment indicates that by incorporating the CoG shift into the model, excellent control results can be achieved.

\noindent\textbf{Ground-contact mode.} In the ground-contact mode, similar to Lai et al. \cite{lai2025trofybot}, we keep the pitch angle $\theta_T$ constant at $0^\circ$ and track an 8-shaped trajectory in order to maximize the energy utilization. 
As shown in Fig.~\ref{rmse}(b), the maximum speed and acceleration reached \qty{3}{m/s} and \qty{2.5}{m/s^2}, respectively, with a trajectory tracking RMSE of \qty{0.074}{m}. 
To compare the stability, we measure the pitch angle from a part of the trajectory, indicated by the green dashed rectangle in Fig.~\ref{rmse}(b).
We plot our results with the results from Zhang et al. \cite{zhang2023model} and Lai et al. \cite{lai2025trofybot} in Fig.~\ref{pitch}(a).
The pitch angle fluctuation of our robot is within a small range from $-4.41^\circ$ to $2.51^\circ$, compared to $-2.2^\circ$ to $9.9^\circ$ for Lai's experiment and a much wider $-12.6^\circ$ to $62.3^\circ$ for Zhang's experiment. 
This result demonstrates that the designed CoG shift, coupled with precise thrust control of our propulsion system, enables a more stable attitude control during high-speed ground travel.

To maximize ground-contact mode efficiency, the robot rotated $90^\circ$ about the $y_t$ axis, reducing ground clearance from \qty{11}{cm} to just \qty{2}{cm}.
The control performance of our propulsion system enables robot to be manipulated at any pitch angle during ground rolling, allowing it to overcome obstacles. 
In Fig.~\ref{rmse}(c), we can see that by controlling $\theta_T$ to improve ground clearance, the robot successfully crosses a ground obstacle with a height of \qty{9}{cm}.

\subsection{Water Domain Experiments}
Underwater vehicles are characterized by highly nonlinear dynamics, significant hydrodynamic damping, and added mass effects, which introduce large inertia and response delays.
In this section, we experimentally verify the robot's underwater mobility and demonstrate that the proposed propulsion system can be effectively applied to underwater vehicles.
A water tank ($5\,\text{m} \times 3\,\text{m} \times 1.6\,\text{m}$) is constructed to simulate the marine environment.

\begin{figure}[t]
    \centering
    \includegraphics[width=\linewidth]{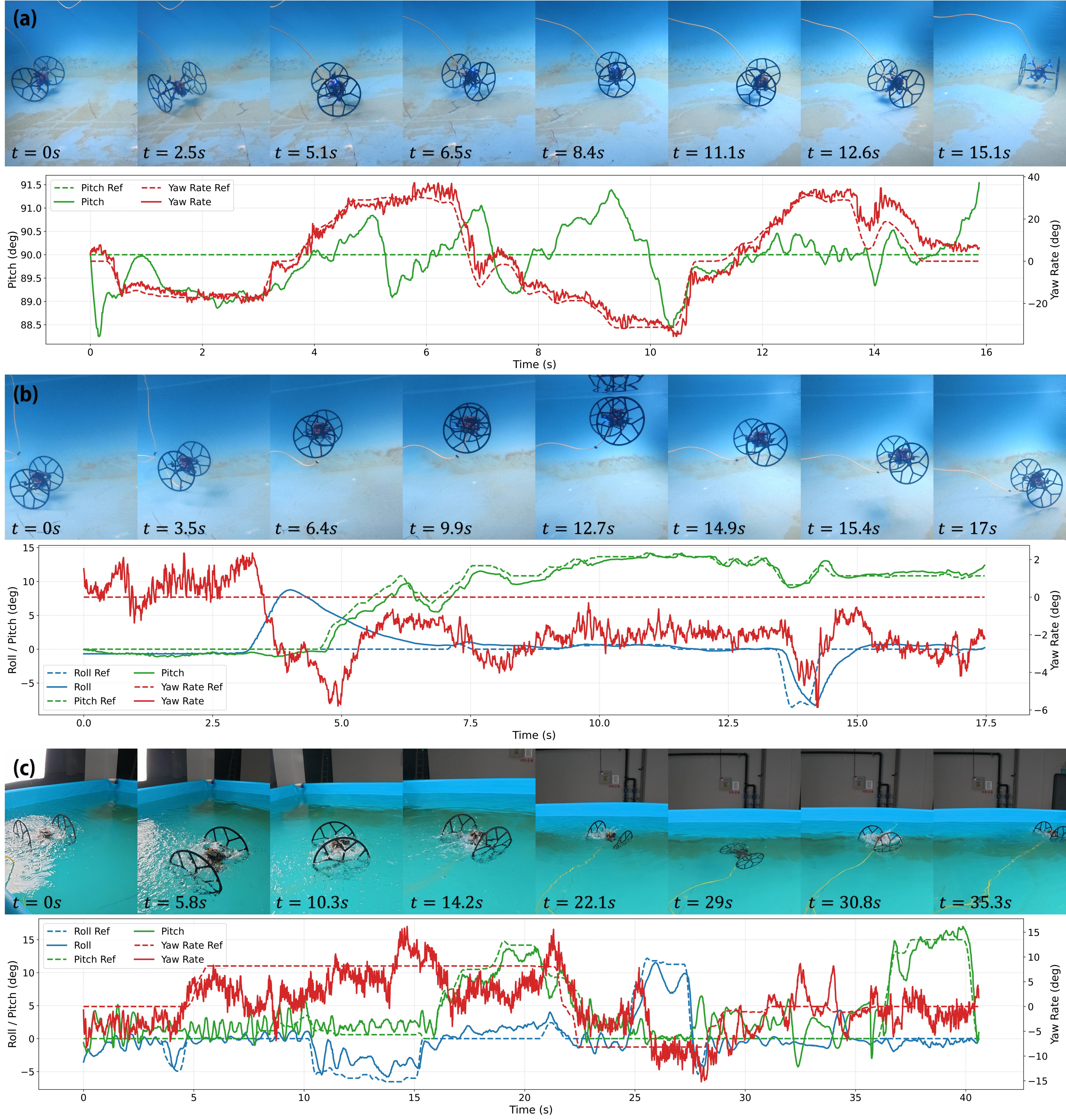}
    \caption{(a) The robot is traveling along an S-shaped path on the seabed. (b) The robot's take-off and landing in water. (c) The robot is moving on the water surface.}
    \label{underwater_test}
\end{figure}

\noindent\textbf{Ground-contact mode.} Fig.~\ref{underwater_test}(a) illustrates the S-shaped trajectory of the robot, demonstrating its good tracking capability of attitude control commands.
Notably, the neutrally buoyant network cable connected to the robot is used exclusively for data acquisition, as the robot is controlled wirelessly.

\noindent\textbf{Non-ground-contact mode.}
Fig.~\ref{underwater_test}(b) and (c) present the submerged cruising and surface movement of \myrobot, respectively.
The results show that the robot effectively tracks control commands despite the influence of the cable.
Compared to the fully submerged state, during the surface-movement phase ($0 < \zeta < 1$), although controllability is maintained, the transition of the medium surrounding the rotors induces small-scale oscillations in the robot's attitude.

\begin{figure}[t]
    \centering
    \includegraphics[width=\linewidth]{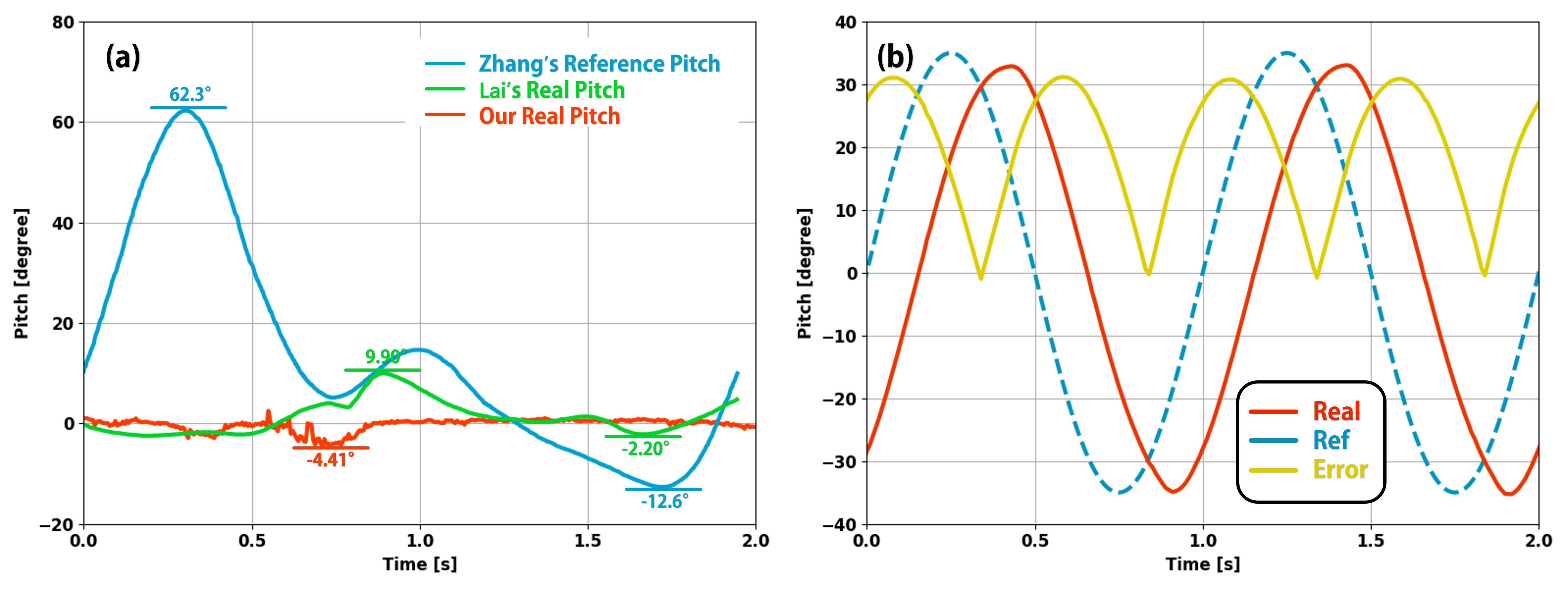}
    \caption{(a) Our real $\theta_T$, TrofyBot's real $\theta_T$ and reference $\theta_T$ of Zhang's reference trajectory from the terrestrial extracted section. (b) The test results of tracking sinusoidal attitude signal and the absolute value of the error.}
    \label{pitch}
\end{figure}

In addition, we conduct a pitch sinusoidal trajectory tracking test, where a sinusoidal signal with a period of $1\,\text{s}$ and an amplitude of $35^\circ$ serves as the target trajectory.
Experimental results are shown in Fig.~\ref{pitch}(b).
The maximum error generated during tracking is $31^\circ$, and the phase lag is only \qty{0.17}{s}.
For comparison, Liu et al.~\cite{liu2024wukong} conduct the same tracking test on the Wukong robot, which adopts a traditional ESC combined with a dedicated underwater motor and propeller, utilizing the incremental nonlinear dynamic inversion (INDI) algorithm to enhance responsiveness.
However, its maximum tracking error exceeds $35^\circ$, and the phase lag is approximately \qty{0.2}{s}. The comparison demonstrates that \myrobot exhibits higher accuracy and faster response in dynamic attitude tracking.

\subsection{Cross-Domain Transition}
To demonstrate the smooth air-land transitions, we first conduct a tracking experiment involving an 8-shaped trajectory in two domains. 
During the mode-switching process, a $90^\circ$ discrepancy in the pitch angle exists between the two modes, and an unsmooth reference signal can lead to system divergence.
Therefore, we pre-plan the pitch angle based on the lift-off altitude; specifically, the robot rotates $\theta_T$ on the ground at a rate of $\dot{\theta} = \pm 60^\circ/\text{s}$ to approach the aerial attitude.
The maximum velocity and acceleration of this trajectory are \qty{2}{m/s} and \qty{2}{m/s^2}, respectively.
The trajectory tracking results are illustrated in Fig.~\ref{rmse}(c), yielding a RMSE of \qty{0.084}{m}.

\begin{figure}[t]
    \centering
    \includegraphics[width=\linewidth]{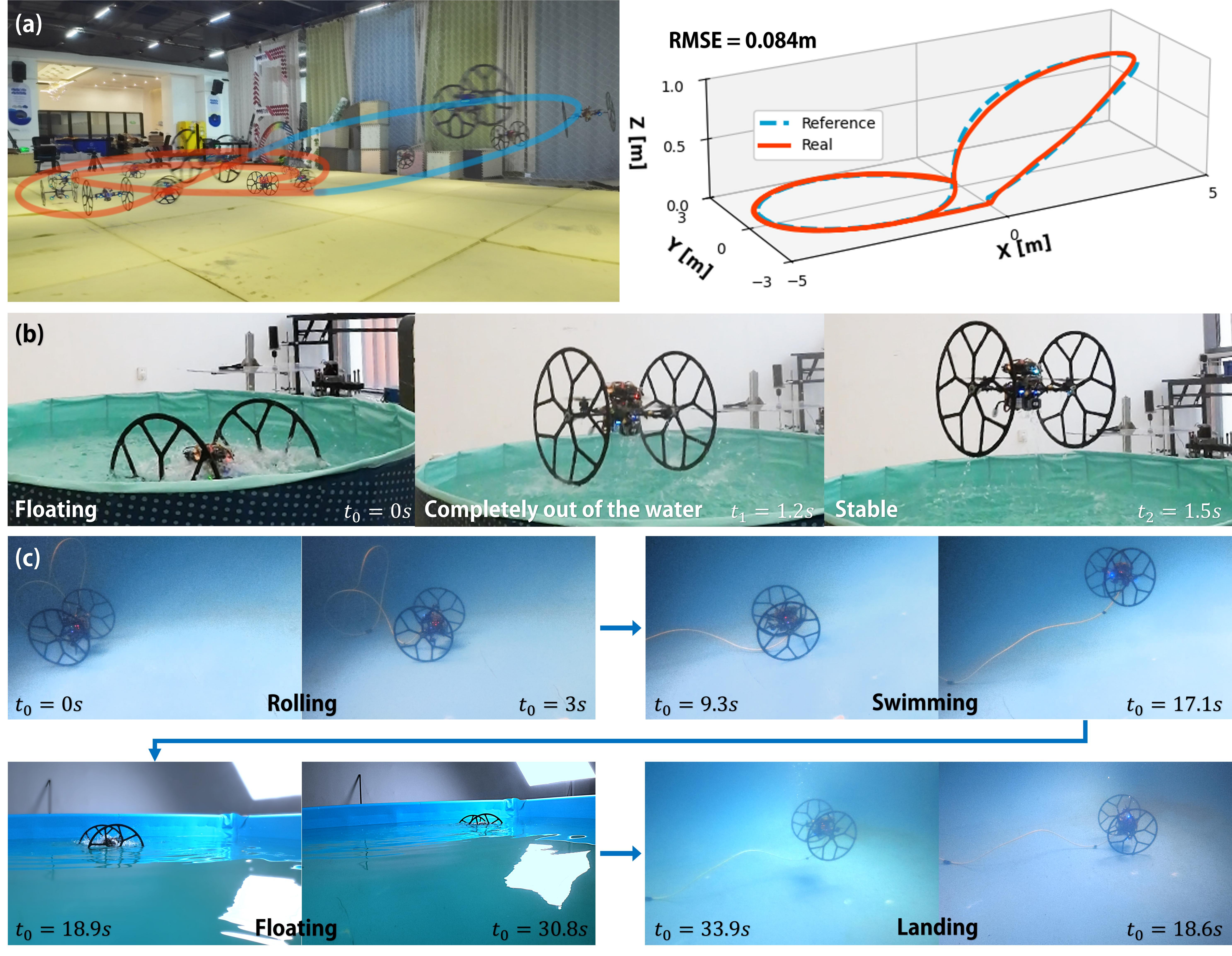}
    \caption{(a) The robot's aerial-terrestrial mode transition is validated by tracking an 8-shaped trajectory, where the reference and the real position are compared. (b) The aquatic-aerial mode conversion of the robot. (c) The modal transformation of the robot in water.}
    \label{transition}
\end{figure}

In air-water transitions, while water entry is achieved by a simple descent, taking off from the water surface poses the primary challenge.
As illustrated in Fig.~\ref{transition}(b), the robot initially floats on the surface. The high torque generated by the FOC propulsion system enables the motors to overcome water drag and rapidly reach the rotational speed required for a swift takeoff.
The robot leaves the water surface in just \qty{1.2}{s}, and once its attitude is stabilized, it can switch to the HNMPC method for autonomous trajectory tracking.

As presented in Fig.~\ref{transition}(c), the robot shows a complete locomotion transition process within the aquatic medium. Specifically, it sequentially performs seabed rolling, submerged cruising, surface movement, and finally descends to seabed.
The above results validate \myrobot's capability for smooth cross-domain transitions and seamless switching of control methods.

\subsection{Performance Comparison with Related Robots}
As shown in Table~\ref{benchmark}, we evaluated \myrobot against representative robots across six key dimensions.
We can see that although our robot lacks the ability to control thrust direction, it demonstrates significant advantages in other key performance indicators.
\begin{table}[h]
    \centering
    \includegraphics[width=\linewidth]{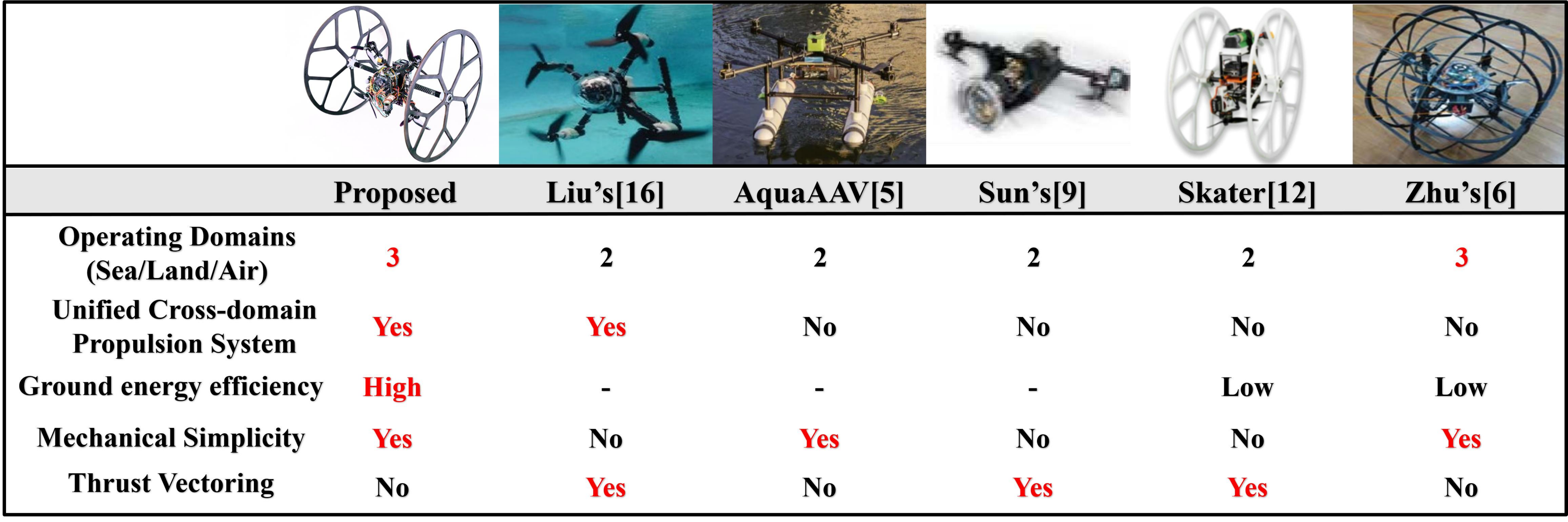}
    \caption{Performance comparison with several robots.}
    \label{benchmark}
\end{table}
\section{Conclusion}\label{CONCLUSION}
This paper presents a novel triphibious robot capable of aerial, terrestrial, and aquatic motion, by a minimalist design combining a quadcopter structure with two passive wheels, without extra actuators.
By utilizing an eccentric CoG design and a unified FOC propulsion system, the robot achieves high efficiency and cross-domain transitions.
An HNMPC-PID controller further ensures stable maneuverability across diverse domains.
In subsequent research, factors such as water flow disturbances should be incorporated, along with more robust cross-domain position controllers, to achieve more accurate and flexible comprehensive working ability across all fields.

\bibliographystyle{unsrt}
\bibliography{article}

 \vspace{-0.5cm}
 
\begin{IEEEbiography}[{\includegraphics
 [width=1in,height=1.25in,clip,
 keepaspectratio]{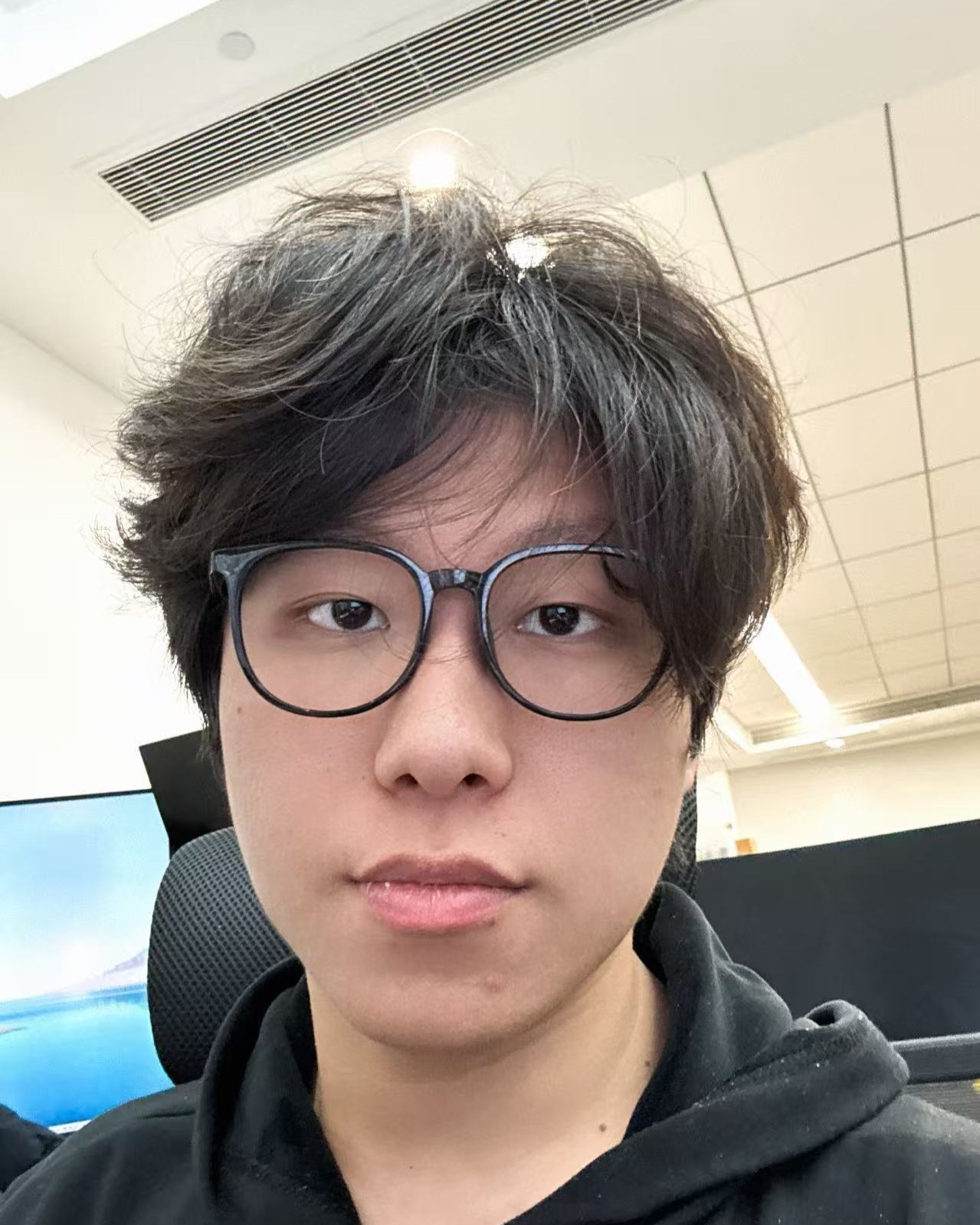}}]
 {Xiangyu Li}received the B.S. degree in electronic information engineering from China Jiliang University, Hangzhou, China, in 2023. He is currently pursuing the M.S. degree in Electronic Information at Zhejiang University, Hangzhou, China. His research interests include multi-domain robot design and control.
 \end{IEEEbiography}
 \vspace{-0.5cm}
 
 \begin{IEEEbiography}[{\includegraphics
 [width=1in,height=1.25in,clip,
 keepaspectratio]{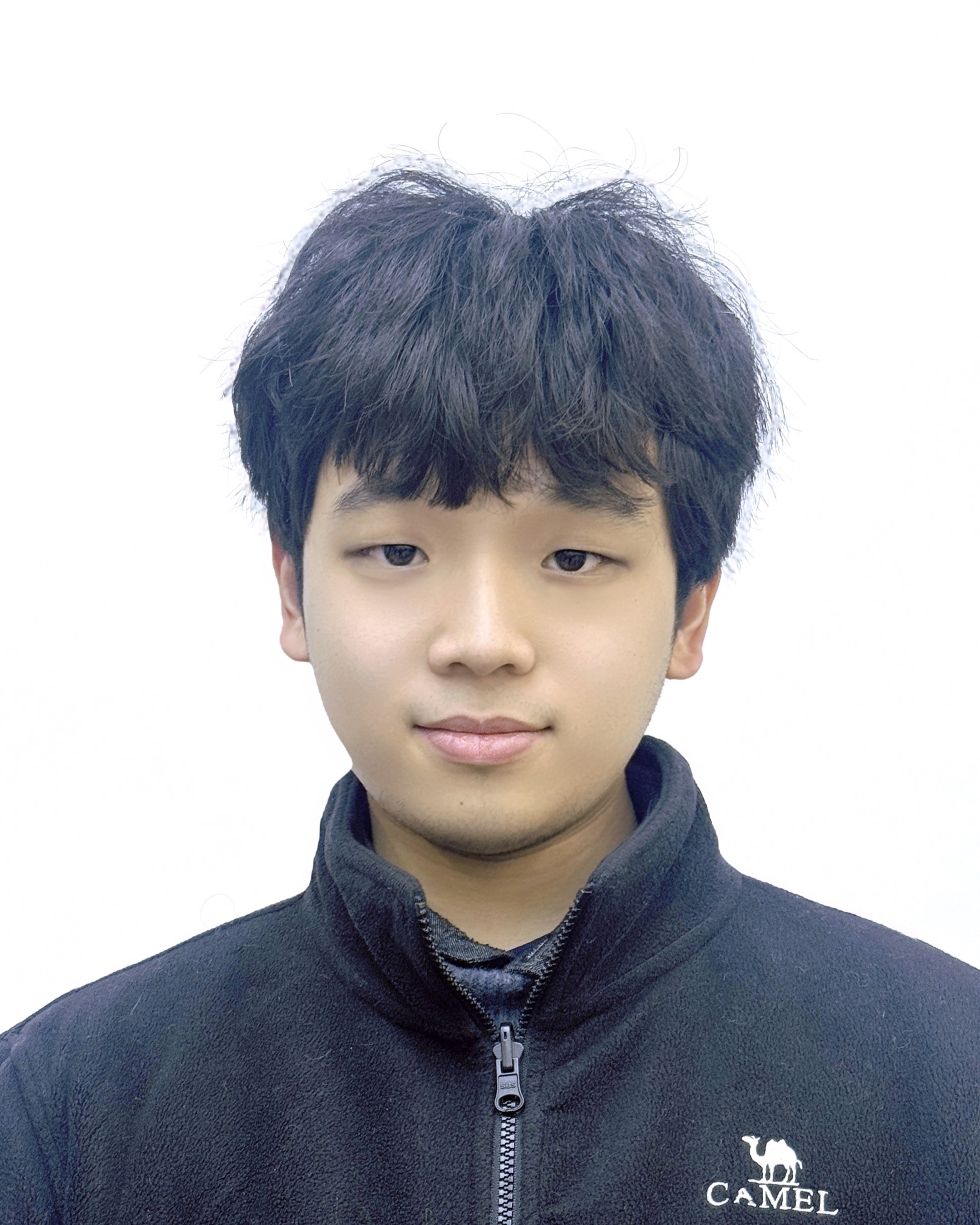}}]
 {Tiancheng Lai}received the B.S. degree in automation and engineering from the Harbin Institute of Technology, Shenzhen, China, in 2024. He is currently pursuing the M.S. degree in control science and engineering in the Fast Lab at Zhejiang University, Zhejiang, China.
His current research interests include motion planning and autonomous exploration of unmanned systems and aerial robots.
 \end{IEEEbiography}

  \vspace{-0.5cm}

\begin{IEEEbiography}[{\includegraphics
 [width=1in,height=1.25in,clip,
 keepaspectratio]{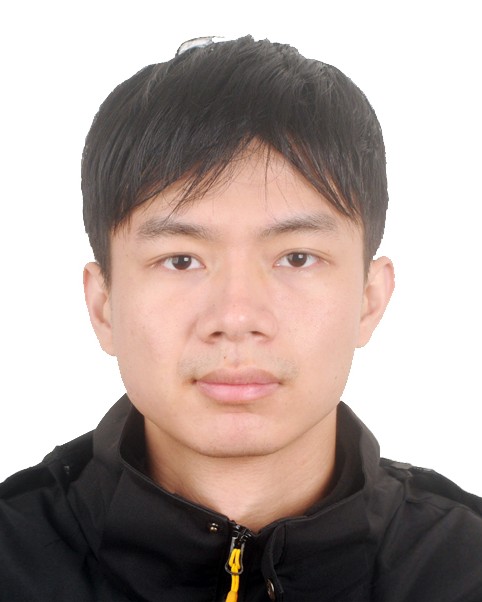}}]
 {Mingwei Lai}received the B.S. degree in Automation from Huazhong University of Science and Technology in 2023. He is currently working toward the M.S. degree in Control Science and Engineering from Zhejiang University. His research interests include motion planning and MPC control.
 \end{IEEEbiography}

 \vspace{-0.5cm}
 
  \begin{IEEEbiography}[{\includegraphics
 [width=1in,height=1.25in,clip,
 keepaspectratio]{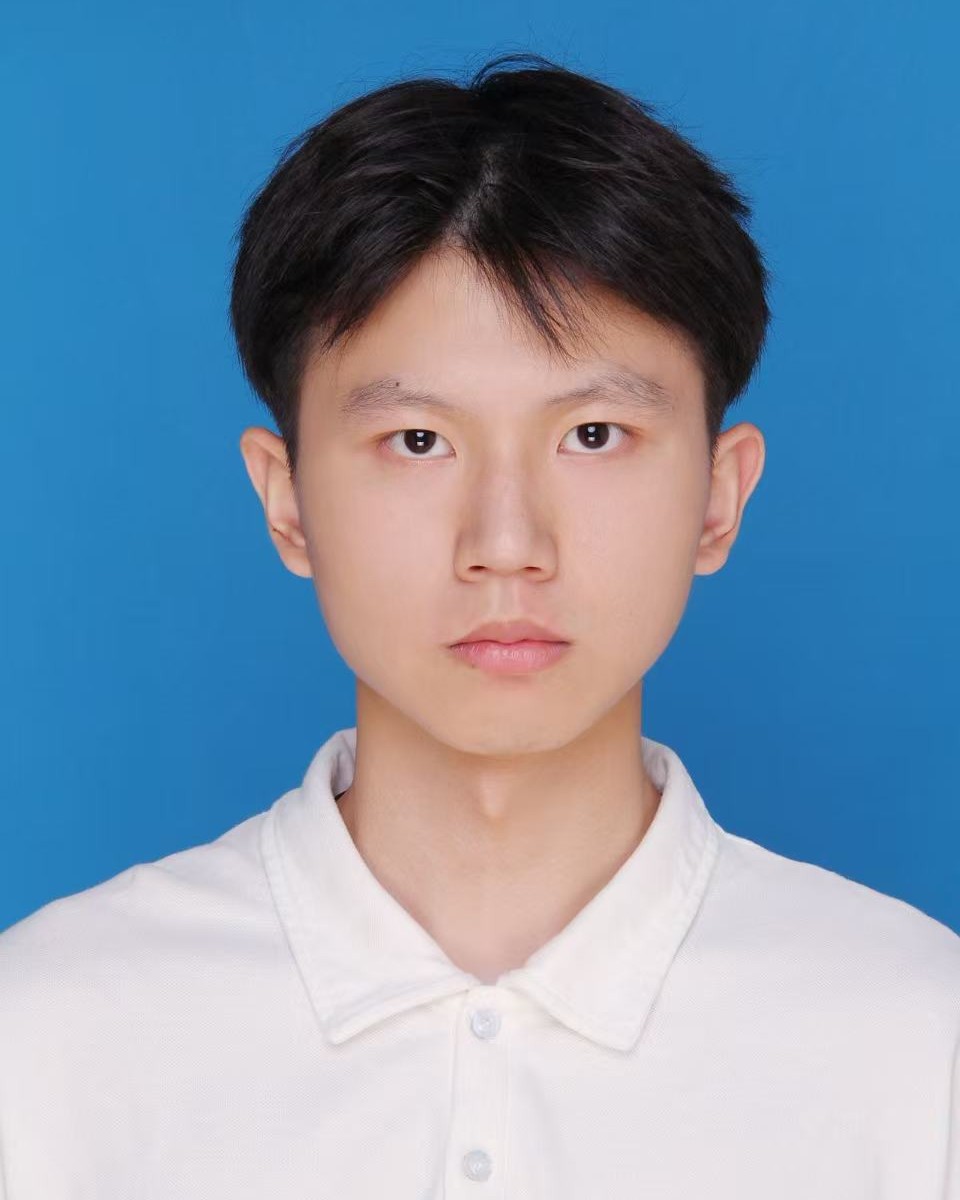}}]
 {Junxiao Lin}(Student Member, IEEE) received the B.Eng. degree in mechanical engineering from Zhejiang University, Hangzhou, China, in 2023, where he is currently pursuing the M.Phil. degree in control engineering.
His research interests include mobile robots, design and control.
 \end{IEEEbiography}
 
 \vspace{-0.5cm}
  \begin{IEEEbiography}[{\includegraphics
 [width=1in,height=1.25in,clip,
 keepaspectratio]{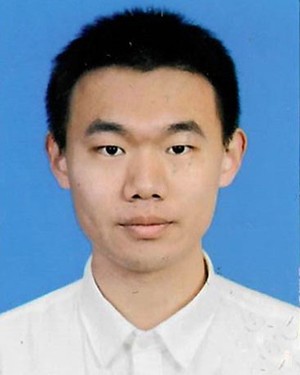}}]
 {Mengke Zhang}received the B.Eng. degree in Automation from Zhejiang University, Hangzhou, China, in 2021. He is currently pursuing the Ph.D. degree at the Fast Lab, Zhejiang University. His research interests include motion planning and uneven terrains.
 \end{IEEEbiography}

 \vspace{-0.5cm}
 
  \begin{IEEEbiography}[{\includegraphics
 [width=1in,height=1.25in,clip,
 keepaspectratio]{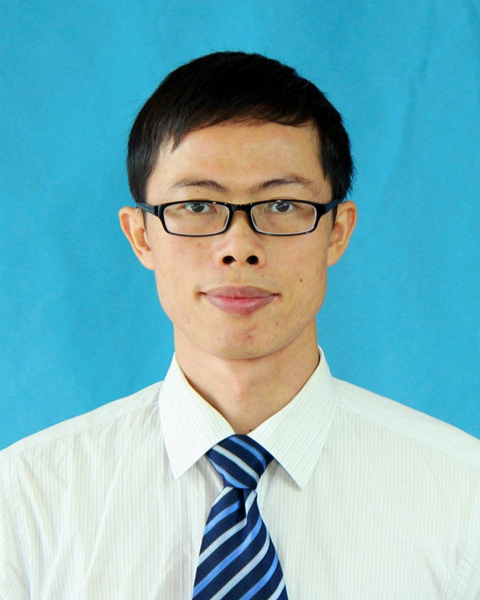}}]
 {Junping Zhi} graduated from Hainan University in 2009 with a Bachelor's degree in Mechanical Design, Manufacturing, and Automation. He is currently engaged in the maintenance and management of automated tobacco machinery. In addition, he is conducting collaborative research with the Zhejiang University Huzhou Institute, focusing on visual imaging technology and the application of unmanned aerial vehicles within the tobacco processing sector.
 \end{IEEEbiography}

 \vspace{-0.5cm}
 
  \begin{IEEEbiography}[{\includegraphics
 [width=1in,height=1.25in,clip,
 keepaspectratio]{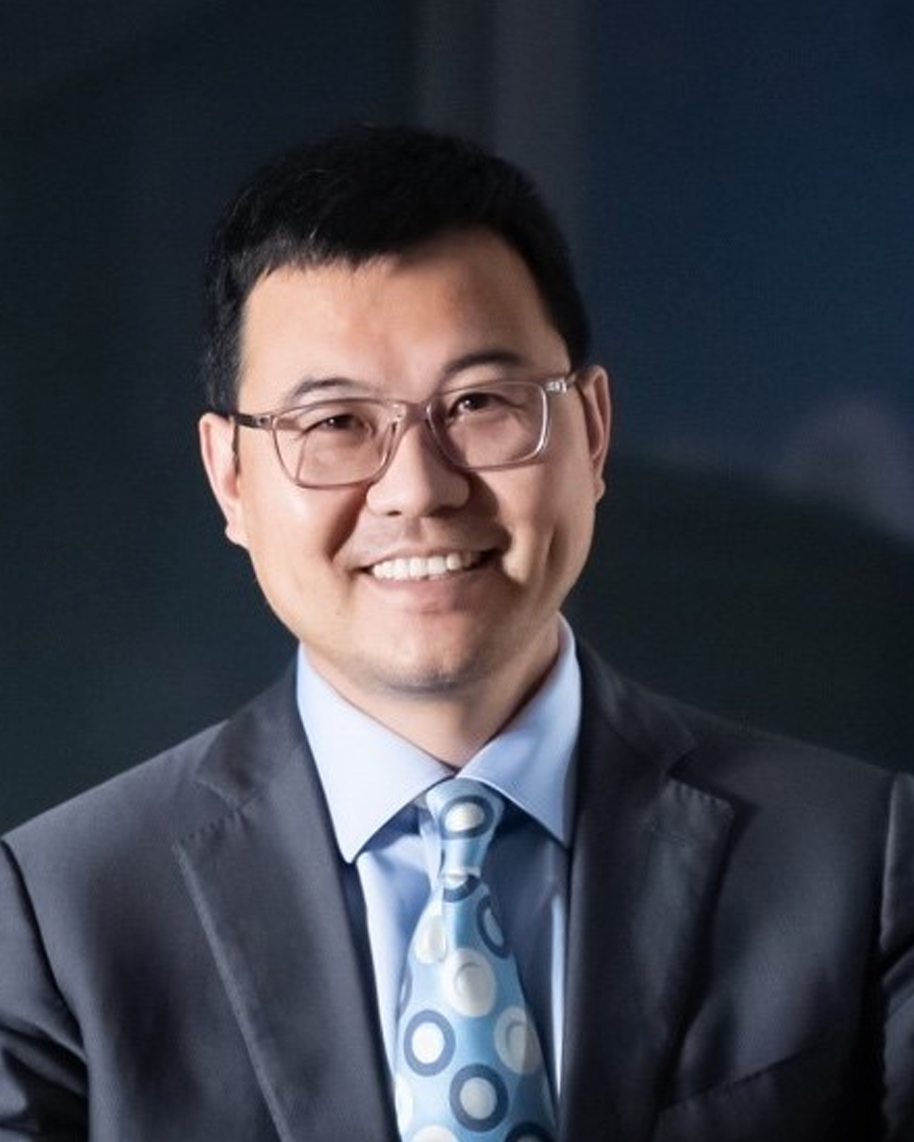}}]
 {Chao Xu}received the Ph.D. degree in mechanical engineering from Lehigh University in 2010. He is currently the Associate Dean and a Professor with the College of Control Science and Engineering, Zhejiang University. He is the Inaugural Dean of ZJU Huzhou Institute. His research expertise is flying robotics and control-theoretic learning. He has published over 100 articles in international journals, including Science Robotics and Nature Machine Intelligence. He will join the organization committee of the IROS-2025 in Hangzhou.
 \end{IEEEbiography}
 
  \vspace{-0.5cm}

  \begin{IEEEbiography}[{\includegraphics
 [width=1in,height=1.25in,clip,
 keepaspectratio]{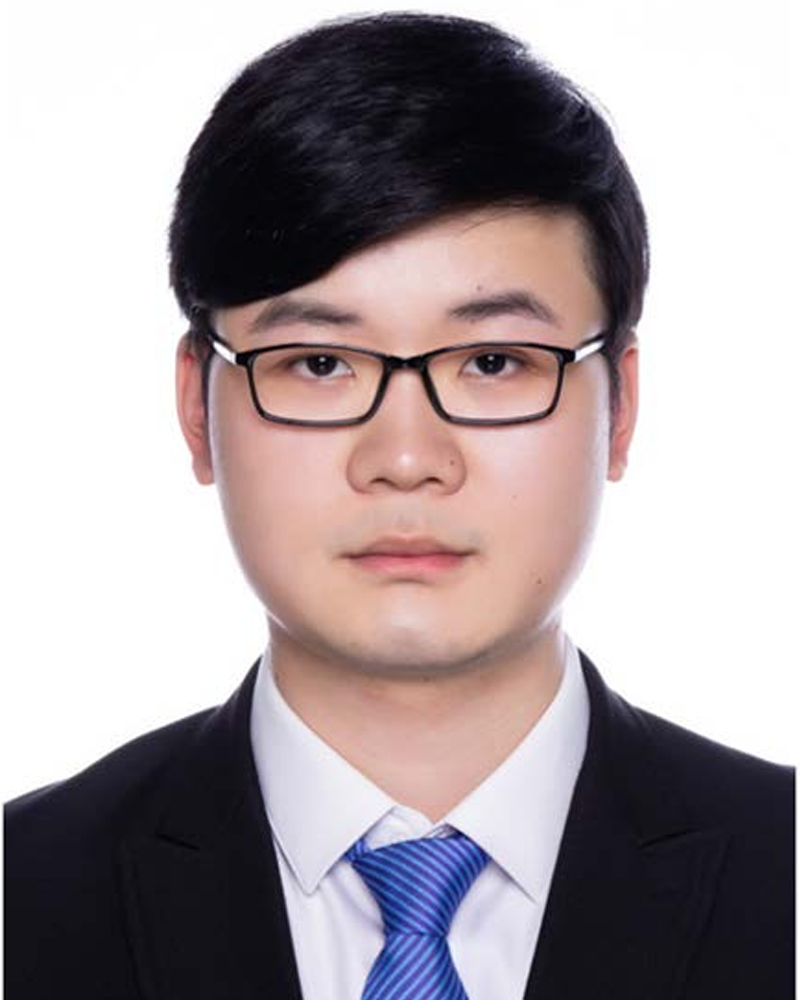}}]
 {Fei Gao}received the Ph.D. degree in electronic and computer engineering from the Hong Kong University of Science and Technology, Hong Kong, in 2019. He is currently a tenured associate professor at the Department of Control Science and Engineering, Zhejiang University. His research interests include aerial robots, autonomous navigation,motion planning, optimization, and localization and mapping.
 \end{IEEEbiography}

 \vspace{-0.5cm}
 
  \begin{IEEEbiography}[{\includegraphics
 [width=1in,height=1.25in,clip,
 keepaspectratio]{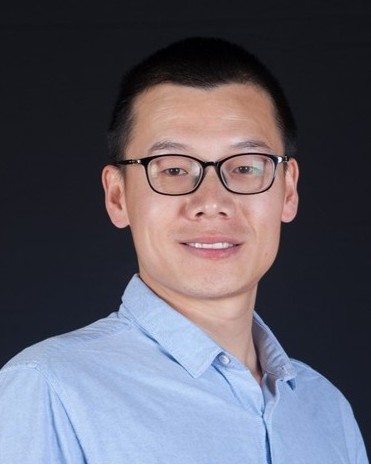}}]
 {Yanjun Cao}received his Ph.D. degree in computer and software engineering from the University of Montreal, Polytechnique Montreal, Canada, in 2020. He is currently an associate researcher at the Huzhou Institute of Zhejiang University, as a PI in the Center of Swarm Navigation. He leads the Field Intelligent Robotics Engineering group of the Field Autonomous System and Computing Lab. His research focuses on key challenges in multi-robot systems, such as collaborative localization, autonomous navigation, perception and communication.
 \end{IEEEbiography}
\end{document}